\documentclass[11pt]{article}

\usepackage[T1]{fontenc}
\usepackage{hyperref}
\usepackage{url}
\usepackage{graphicx} 
\usepackage{subfigure} 
\usepackage{bm} 
\usepackage{amsmath}
\usepackage{natbib} 
\usepackage{multirow}

\usepackage[a4paper, total={6in, 8in}]{geometry}

\title{Sigmoid-Weighted Linear Units for Neural Network Function
  Approximation in Reinforcement Learning}

\author{
Stefan Elfwing$^{\rm a}$ \\
\texttt{elfwing@atr.jp}
\and
Eiji Uchibe$^{\rm a,b}$ \\
\texttt{uchibe@atr.jp} 
\and
Kenji Doya$^{\rm b}$ \\
\texttt{doya@oist.jp} 
}

\date{$^{\rm a}$Dept. of Brain Robot Interface, ATR Computational
  Neuroscience Laboratories, 2-2-2 Hikaridai, Seikacho, Soraku-gun,
  Kyoto 619-0288, Japan\\ 
  $^{\rm b}$Neural Computation Unit, Okinawa
  Institute of Science and Technology Graduate University, 1919-1
  Tancha, Onna-son, Okinawa 904-0495, Japan\\ }

\begin{document}

\maketitle

\begin{abstract}
\noindent In recent years, neural networks have enjoyed a renaissance
as function approximators in reinforcement learning. Two decades after
Tesauro's TD-Gammon achieved near top-level human performance in
backgammon, the deep reinforcement learning algorithm DQN achieved
human-level performance in many Atari 2600 games. The purpose of this
study is twofold. First, we propose two activation functions for
neural network function approximation in reinforcement learning: the
sigmoid-weighted linear unit (SiLU) and its derivative function
(dSiLU). The activation of the SiLU is computed by the sigmoid
function multiplied by its input. Second, we suggest that the more
traditional approach of using on-policy learning with eligibility
traces, instead of experience replay, and softmax action selection
with simple annealing can be competitive with DQN, without the need
for a separate target network. We validate our proposed approach by,
first, achieving new state-of-the-art results in both stochastic
SZ-Tetris and Tetris with a small 10$\times$10 board, using
TD($\lambda$) learning and shallow dSiLU network agents, and, then, by
outperforming DQN in the Atari 2600 domain by using a deep
Sarsa($\lambda$) agent with SiLU and dSiLU hidden units.
\end{abstract}

\section{Introduction}
Neural networks have enjoyed a renaissance as function approximators
in reinforcement learning~\citep{Sutton98} in recent years. The DQN
algorithm~\citep{Mnih15}, which combines Q-learning with a deep neural
network, experience replay, and a separate target network, achieved
human-level performance in many Atari 2600 games. Since the
development of the DQN algorithm, there have been several proposed
improvements, both to DQN specifically and deep reinforcement learning
in general. \Citet{Hasselt15} proposed double DQN to reduce
overestimation of the action values in DQN and \citet{Schaul2016}
developed a framework for more efficient replay by prioritizing
experiences of more important state transitions. \citet{Wang2016}
proposed the dueling network architecture for more efficient learning
of the action value function by separately estimating the state value
function and the advantages of each action. \citet{Mnih2016} proposed
a framework for asynchronous learning by multiple agents in parallel,
both for value-based and actor-critic methods.

The purpose of this study is twofold. First, motivated by the high
performance of the expected energy restricted Boltzmann machine
(EE-RBM) in our earlier studies \citep{Elfwing15,Elfwing16}, we
propose two activation functions for neural network function
approximation in reinforcement learning: the sigmoid-weighted linear
unit (SiLU) and its derivative function (dSiLU). The activation of the
SiLU is computed by the sigmoid function multiplied by its input and
it looks like a continuous and ``undershooting'' version of the linear
rectifier unit (ReLU)~\citep{Hahnloser00}. The activation of the dSiLU
looks like steeper and ``overshooting'' version of the sigmoid
function.

Second, we suggest that the more traditional approach of using
on-policy learning with eligibility traces, instead of experience
replay, and softmax action selection with simple annealing can be
competitive with DQN, without the need for a separate target
network. Our approach is something of a throwback to the approach used
by~\citet{Tesauro1994} to develop TD-Gammon more than two decades
ago. Using a neural network function approximator and TD($\lambda$)
learning~\citep{Sutton88}, TD-Gammon reached near top-level human
performance in backgammon, which to this day remains one of the most
impressive applications of reinforcement learning.

To evaluate our proposed approach, we first test the performance of
shallow network agents with SiLU, ReLU, dSiLU, and sigmoid hidden units in
stochastic SZ-Tetris, which is a simplified but difficult version of
Tetris. The best agent, the dSiLU network agent, improves the average
state-of-the-art score by 20\,\%. In stochastic SZ-Tetris, we also
train deep network agents using raw board configurations as states. An
agent with SiLUs in the convolutional layers and dSiLUs in the
fully-connected layer (SiLU-dSiLU) outperforms the previous
state-of-the-art average final score. We thereafter train a dSiLU
network agent in standard Tetris with a smaller, 10$\times$10, board
size, achieving a state-of-the-art score in this more competitive
version of Tetris as well. We then test a deep SiLU-dSiLU network
agent in the Atari 2600 domain. It improves the mean DQN normalized
scores achieved by DQN and double DQN by 232\,\% and 161\,\%,
respectively, in 12 unbiasedly selected games. We finally analyze the
ability of on-policy value-based reinforcement learning to accurately
estimate the expected discounted returns and the importance of softmax
action selection for the games where our proposed agents performed
particularly well.

\section{Method}
\subsection[TD(lambda) and Sarsa(lambda)]{TD($\lambda$) and Sarsa($\lambda$)}
In this study, we use two reinforcement learning algorithms:
TD($\lambda$)~\citep{Sutton88} and
Sarsa($\lambda$)~\citep{Rummery94,Sutton96}. TD($\lambda$) learns an
estimate of the state-value function, $V^{^{\pi}}$, and
Sarsa($\lambda$) learns an estimate of the action-value function,
$Q^{^{\pi}}$, while the agent follows policy $\pi$. If the
approximated value functions, $V_t\approx V^{^{\pi}}$ and $Q_t\approx
Q^{^{\pi}}$, are parameterized by the parameter vector
$\boldsymbol{\theta}_t$, then the gradient-descent learning update of
the parameters is computed by
\begin{equation}
  \boldsymbol{\theta}_{t+1} = \boldsymbol{\theta}_t + \alpha\delta_t\boldsymbol{e}_t,
  \label{eq:upd}
\end{equation}
where the TD-error, $\delta_t$, is
\begin{equation}
\delta_t = r_{t} +\gamma V_t(s_{t+1}) - V_t(s_{t})
\end{equation}
for TD($\lambda$) and 
\begin{equation}
\delta_t = r_{t} +\gamma Q_t(s_{t+1},a_{t+1}) - Q_t(s_{t},a_{t})
\end{equation}
for Sarsa($\lambda$). The eligibility trace vector,
$\boldsymbol{e}_t$, is
\begin{equation}
\boldsymbol{e}_t = \gamma\lambda \boldsymbol{e}_{t-1} + \nabla_{\boldsymbol{\theta}_t} V_t(s_{t}), \ \boldsymbol{e}_0 = \boldsymbol{0}, 
\label{eq:tracesV}
\end{equation}
for TD($\lambda$) and
\begin{equation}
\boldsymbol{e}_t = \gamma\lambda \boldsymbol{e}_{t-1} + \nabla_{\boldsymbol{\theta}_t} Q_t(s_{t},a_{t}), \ \boldsymbol{e}_0 = \boldsymbol{0},
\label{eq:tracesQ}
\end{equation}
for Sarsa($\lambda$). Here, $s_t$ is the state at time $t$, $a_t$ is
the action selected at time $t$, $r_t$ is the reward for taking action
$a_t$ in state $s_t$, $\alpha$ is the learning rate, $\gamma$ is
the discount factor of future rewards, $\lambda$ is the trace-decay
rate, and $\nabla_{\boldsymbol{\theta}_t} V_t$ and
$\nabla_{\boldsymbol{\theta}_t} Q_t$ are the vectors of partial
derivatives of the function approximators with respect to each
component of $\boldsymbol{\theta}_t$.

\subsection{Sigmoid-weighted Linear Units}
In our earlier work~\citep{Elfwing16}, we proposed the EE-RBM as a
function approximator in reinforcement learning. In the case of state-value based learning,
given a state vector $\mathbf{s}$, an EE-RBM approximates the
state-value function $V$ by the negative expected energy of an
RBM~\citep{Smolensky86,Freund92,Hinton02} network:
\begin{eqnarray}
V(\mathbf{s}) &=& \sum_k z_{k}\sigma(z_{k}) + \sum_ib_is_i, \label{EE-RBM}\\
 z_{k}      &=&   \sum_iw_{ik}s_i + b_k, \\
 \sigma(x) &=& \frac{1}{1+e^{-x}}. 
\end{eqnarray}
Here, $z_{k}$ is the input to hidden unit $k$, $\sigma(\cdot)$ is the
sigmoid function, $b_i$ is the bias weight for input unit $s_i$,
$w_{ik}$ is the weight connecting state $s_i$ and hidden unit $k$,
and $b_k$ is the bias weight for hidden unit $k$. Note that
Equation~\ref{EE-RBM} can be regarded as the output of a one-hidden
layer feedforward neural network with hidden unit activations computed
by $z_{k}\sigma(z_{k})$ and with uniform output weights of one. 

In this study, motivated by the high performance of the EE-RBM in both
the classification~\citep{Elfwing15} and the reinforcement
learning~\citep{Elfwing16} domains, we propose the SiLU as an
activation function for neural network function approximation in
reinforcement learning. The activation $a_k$ of a SiLU $k$ for an
input vector $\mathbf{s}$ is computed by the sigmoid function
multiplied by its input:
\begin{equation}
a_k(\mathbf{s}) = z_k\sigma(z_k).
\end{equation}
\begin{figure}[!thb]
   \begin{center}
    \subfigure
    {
      \includegraphics[width=0.47\textwidth]{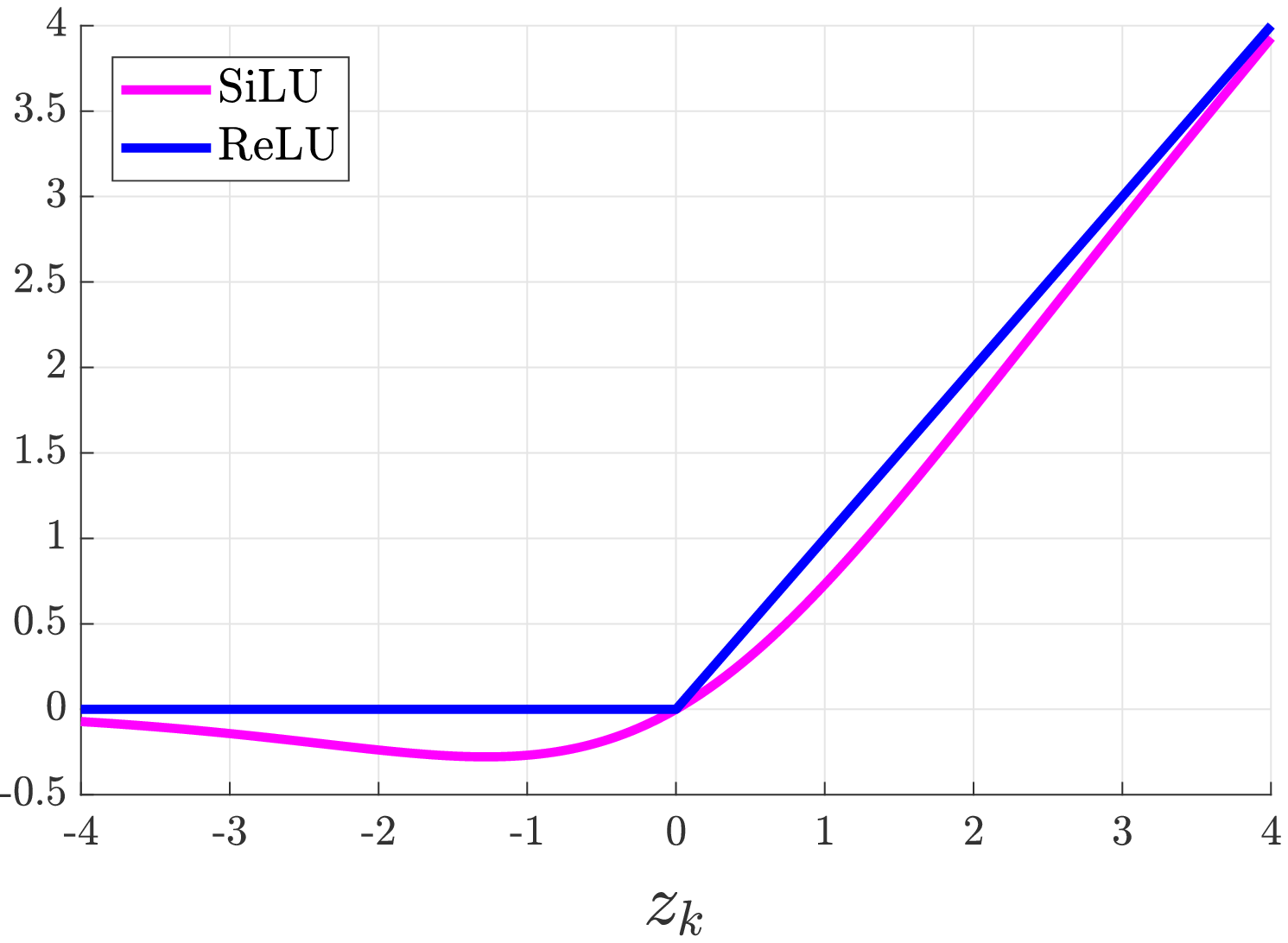} 
    }
    \subfigure
    {
      \includegraphics[width=0.47\textwidth]{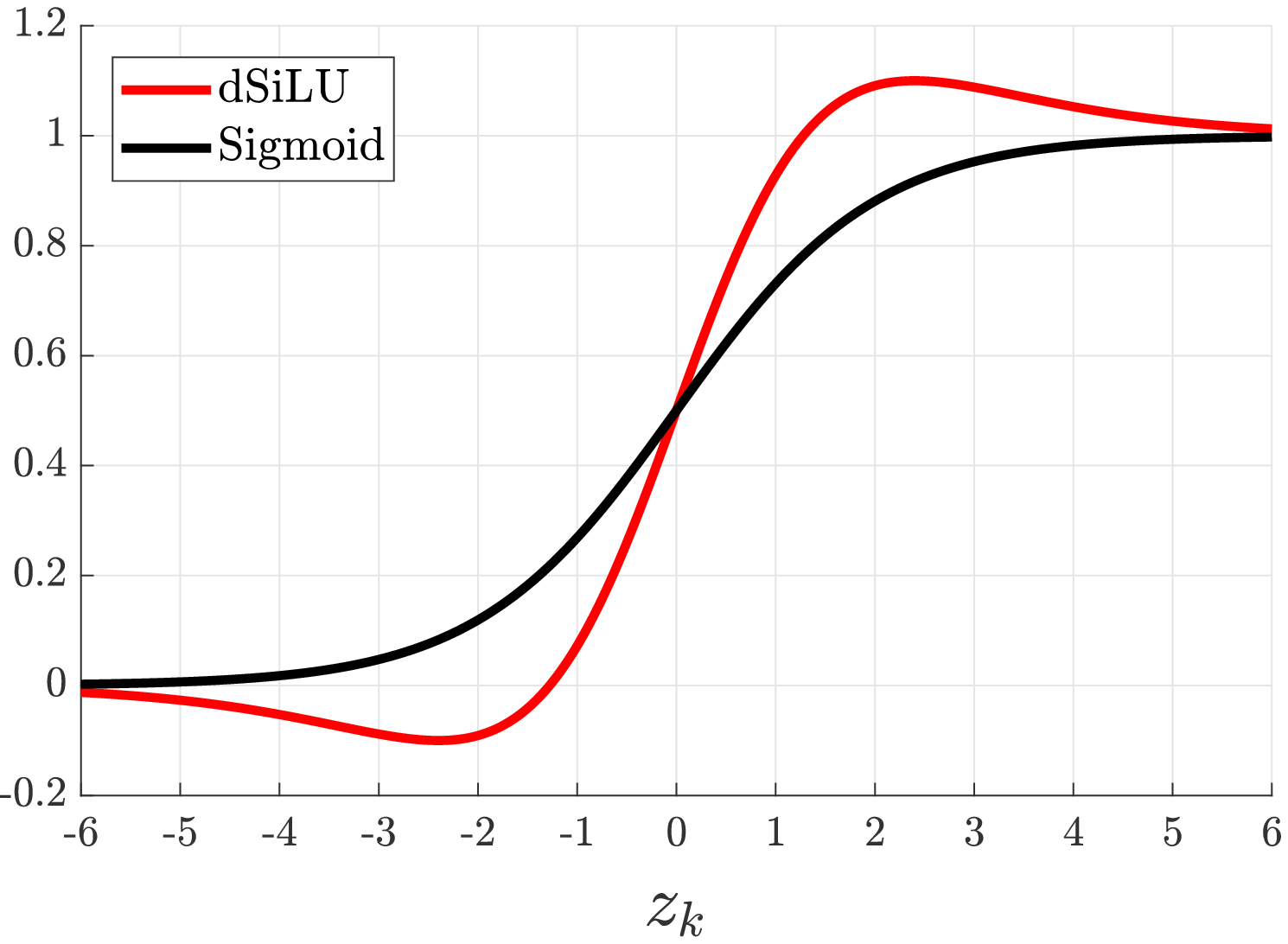} 
    } 
  \end{center}
  \caption{The activation functions of the SiLU and the ReLU (left
    panel), and the dSiLU and the sigmoid unit (right panel).}
  \label{fig:act_units} 
\end{figure}

For $z_k$-values of large magnitude, the activation of the SiLU is
approximately equal to the activation of the ReLU (see left panel in
Figure~\ref{fig:act_units}), i.e., the activation is approximately
equal to zero for large negative $z_k$-values and approximately equal
to $z_k$ for large positive $z_k$-values. Unlike the ReLU (and other
commonly used activation units such as sigmoid and tanh units), the
activation of the SiLU is not monotonically increasing. Instead, it
has a global minimum value of approximately $-0.28$ for $z_{k} \approx
-1.28$. An attractive feature of the SiLU is that it has a
self-stabilizing property, which we demonstrated experimentally in
\citet{Elfwing15}. The global minimum, where the derivative is zero,
functions as a ``soft floor'' on the weights that serves as an
implicit regularizer that inhibits the learning of weights of large
magnitudes.

We propose an additional activation function for neural network
function approximation: the dSiLU. The activation of the dSiLU
is computed by the derivative of the SiLU:
\begin{equation}
a_k(\mathbf{s}) = \sigma(z_k)\left(1 + z_k(1 - \sigma(z_k))\right).
\end{equation}
The activation function of the dSiLU looks like an steeper and
``overshooting'' sigmoid function (see right panel in
Figure~\ref{fig:act_units}). The dSiLU has a maximum value
of approximately $1.1$ and a minimum value of approximately $-0.1$ for
$z_{k} \approx \pm 2.4$, i.e., the solutions to the equation $z_{k} =
-\log\left((z_{k} - 2)/(z_{k} + 2)\right)$.

The derivative of the activation function of the SiLU, used for
gradient-descent learning updates of the neural network weight
parameters (see Equations \ref{eq:tracesV} and \ref{eq:tracesQ}), is
given by
\begin{equation}
\nabla_{w_{ik}} a_k(\mathbf{s}) = \sigma(z_k)\left(1 + z_{k}(1 -
\sigma(z_k))\right)s_i,
\end{equation}
and the derivative of the activation function of the dSiLU is
given by
\begin{equation}
\nabla_{w_{ik}} a_k(\mathbf{s}) = \sigma(z_k)(1-\sigma(z_k))(2 + z_{k}(1 - \sigma(z_k))- z_{k}\sigma(z_k))s_i.
\end{equation}

\subsection{Action selection}
We use softmax action selection with a Boltzmann distribution in all
experiments. For Sarsa($\lambda$), the probability to select action
$a$ in state $s$ is defined as
\begin{equation}
\pi(a | s) = \frac{\exp({Q(s,a)/\tau})}{\sum_b \exp({Q(s,b)/\tau})}.
\end{equation}
For the model-based TD($\lambda$) algorithm, we select an action $a$
in state $s$ that leads to the next state $s'$ with a probability
defined as
\begin{equation}
\pi(a | s) = \frac{\exp({V(f(s,a))/\tau})}{\sum_{b} \exp({V(f(s,b))/\tau})}.
\end{equation}
Here, $f(s,a)$ returns the next state $s'$ according to the state
transition dynamics and $\tau$ is the temperature that controls the
trade-off between exploration and exploitation. We used hyperbolic
annealing of the temperature and the temperature was decreased after
every episode $i$:
\begin{equation}
\tau(i) = \frac{\tau_0}{1 + \tau_ki}.
\label{eq:tau}
\end{equation}
Here, $\tau_0$ is the initial temperature and $\tau_k$ controls the
rate of annealing.

\section{Experiments}
\subsection{SZ-Tetris}
\citet{Szita10} proposed stochastic SZ-Tetris~\citep{Burgiel97} as a
benchmark for reinforcement learning that preserves the core
challenges of standard Tetris but allows faster evaluation of
different strategies due to shorter episodes by removing easier
tetrominoes. Stochastic SZ-Tetris is played on a board of standard
Tetris size with a width of 10 and a height of 20. In each time step,
either an S-shaped tetromino or a Z-shaped tetromino appears with
equal probability. The agent selects a rotation (lying or standing)
and a horizontal position within the board. In total, there are 17
possible actions for each tetromino (9 standing and 8 lying horizontal
positions). After the action selection, the tetromino drops down the
board, stopping when it hits another tetromino or the bottom of the
board. If a row is completed, then it disappears. The agent gets a
score of +1 point for each completed row. An episode ends when a
tetromino does not fit within the board.

For an alternating sequence of S-shaped and Z-shaped tetrominoes, the
upper bound on the episode length in SZ-Tetris is 69\,600 fallen
pieces~\citep{Burgiel97} (corresponding to a score of 27\,840 points),
but the maximum episode length is probably much shorter, maybe a few
thousands~\citep{Szita10}. That means that to evaluate a good
strategy SZ-Tetris requires at least five orders of magnitude less
computation than standard Tetris.

The standard learning approach for Tetris has been to use a
model-based setting and define the evaluation function or state-value
function as the linear combination of hand-coded features. Value-based
reinforcement learning algorithms have a lousy track record using this
approach. In regular Tetris, their reported performance levels are
many magnitudes lower than black-box methods such as the cross-entropy
(CE) method and evolutionary approaches.  In stochastic SZ-Tetris, the
reported scores for a wide variety of reinforcement learning
algorithms are either approximately zero~\citep{Szita10} or in the
single
digits~\footnote{http://barbados2011.rl-community.org/program/SzitaTalk.pdf}.

Value-based reinforcement learning has had better success in
stochastic SZ-Tetris when using non-linear neural network based
function approximators. \citet{Fausser13} achieved a score of about
130 points using a shallow neural network function approximator with
sigmoid hidden units. They improved the result to about 150 points by
using an ensemble approach consisting of ten neural networks. We
achieved an average score of about 200 points using three different
neural network function approximators: an EE-RBM, a free energy RBM,
and a standard neural network with sigmoid hidden
units~\citep{Elfwing16}. \citet{Jaskowski15} achieved the current
state-of-the-art results using systematic n-tuple networks as function
approximators: average scores of 220 and 218 points achieved by the
evolutionary VD-CMA-ES method and TD-learning, respectively, and the
best mean score in a single run of 295 points achieved by TD-learning.

In this study, we compare the performance of different hidden
activation units in two learning settings: 1) shallow network agents
with one hidden layer using hand-coded state features and 2) deep
network agents using raw board configurations as states, i.e., a state
node is set to one if the corresponding board cell is occupied by a
tetromino and set to zero otherwise.

\begin{figure}[!thb]
   \begin{center}
    \subfigure
    {
      \includegraphics[width=0.47\textwidth]{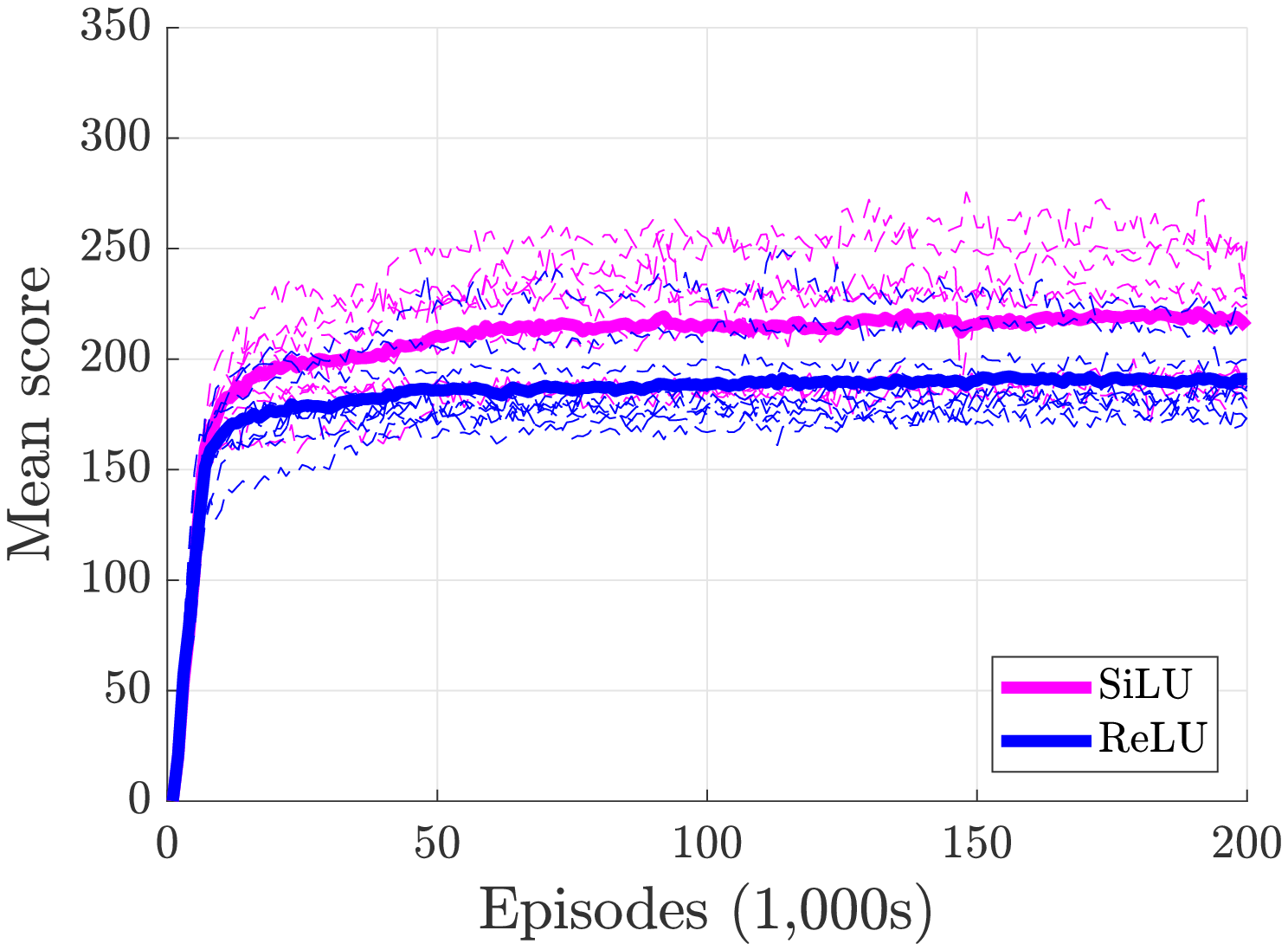} 
    }
    \subfigure
    {
     \includegraphics[width=0.47\textwidth]{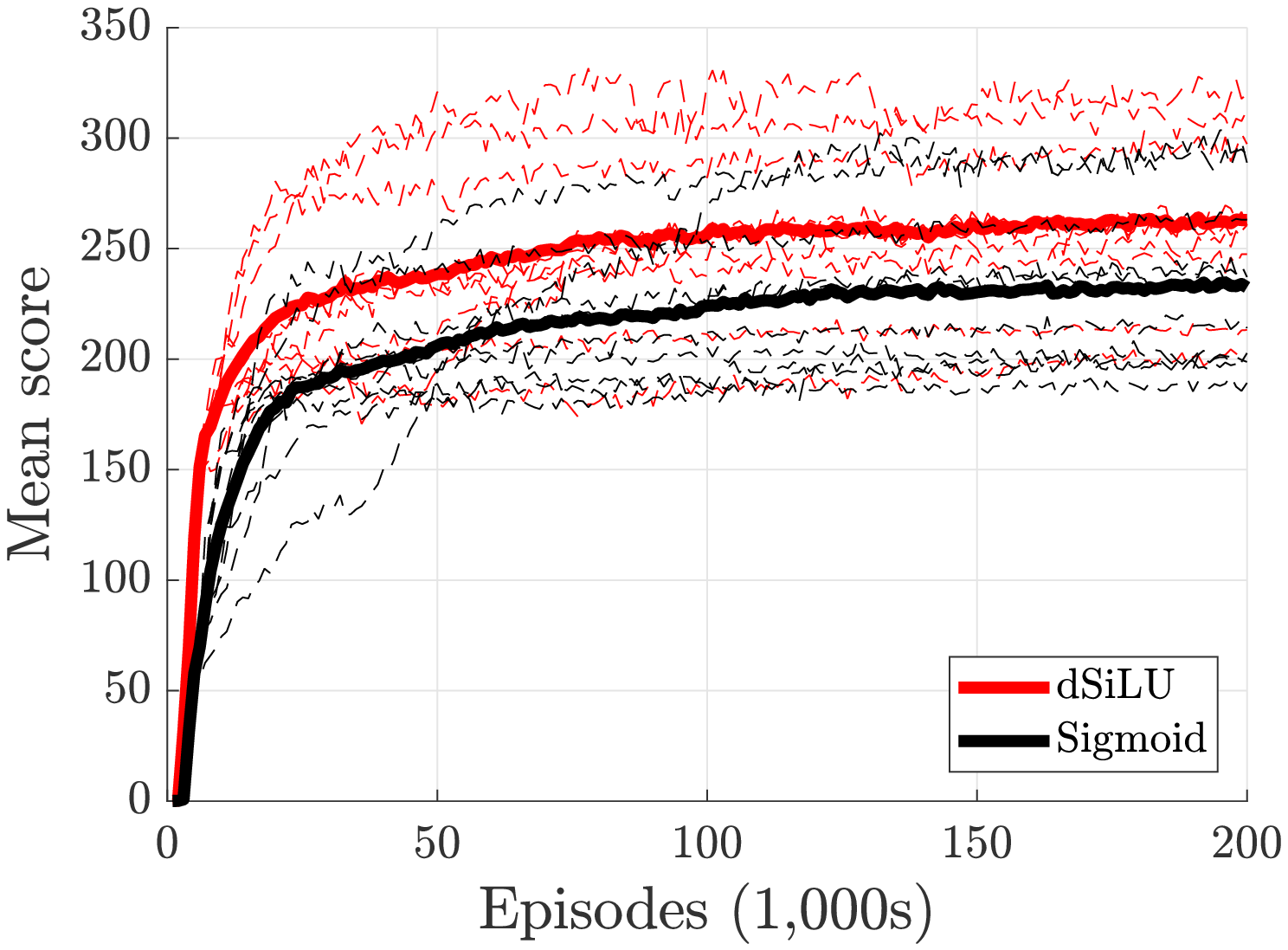} 
    } 
  \end{center}
  \caption{Learning curves in stochastic SZ-Tetris for the four types
    of shallow neural network agents. The figure shows the average scores 
    over ten separate runs (tick solid lines) and the scores of individual
    runs (thin dashed lines). The mean scores were computed over every
    1,000 episodes.}
  \label{fig:res_SZtet} 
\end{figure}

In the setting with state features, we trained shallow network agents
with SiLU, ReLU, dSiLU, and sigmoid hidden units, using the
TD($\lambda$) algorithm and softmax action selection. We used the same
experimental setup as used in our earlier work \citep{Elfwing16}. The
networks consisted of one hidden layer with 50 hidden units and a
linear output layer. The features were similar to the original 21
features proposed by~\citet{Bertsekas96}, except for not including the
maximum column height and using the differences in column heights
instead of the absolute differences. The length of the binary state
vector was 460. The shallow network agents were trained for 200,000
episodes and the experiments were repeated for ten separate runs for
each type of activation unit.

In the deep reinforcement learning setting, we used a deep network
architecture consisting of two convolutional layers with 15 and 50
filters of size $5\times5$ using a stride of 1, a fully-connected
layer with 250 units, and a linear output layer. Both convolutional
layers were followed by max-pooling layers with pooling windows of
size $3\times3$ using a stride of 2. The deep network agents were also
trained using the TD($\lambda$) algorithm and softmax action
selection. We trained three types of deep networks with: 1) SiLUs in
both the convolutional and fully-connected layers (SiLU-SiLU); 2)
ReLUs in both the convolutional and fully-connected layers
(ReLU-ReLU); and 3) SiLUs in the convolutional layers and dSiLUs in
the fully-connected layer (SiLU-dSiLU). The deep network agents were
trained for 200,000 episodes and the experiments were repeated for
five separate runs for each type of network.

\begin{figure}[!thb]
   \begin{center}
     \includegraphics[width=0.48\textwidth]{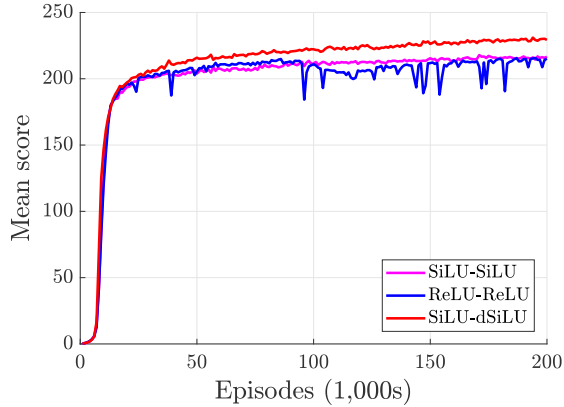} 
  \end{center}
  \caption{Average Learning curves in stochastic SZ-Tetris for the three
    types of deep neural network agents. The figure shows the average scores 
    over five separate runs, computed over every 1,000 episodes.}
  \label{fig:res_deepSZtet} 
\end{figure}

We used the following reward function (proposed by \citet{Fausser13}):
\begin{equation}
r(\mathbf{s}) = e^{-(\textrm{number of holes in }\mathbf{s})/33}.
\label{eq:sz_tetris_r}
\end{equation}
We set $\gamma$ to $0.99$, $\lambda$ to $0.55$, $\tau_0$ to $0.5$, and
$\tau_k$ to $0.00025$. We used a rough grid-like search to find
appropriate values of the learning rate $\alpha$ and it was determined
to be $0.001$ for the four shallow network agents and $0.0001$ for the
three deep network agents.

\begin{table}[!htb]
\caption{Average scores ($\pm$ standard deviations) achieved in
  stochastic SZ-Tetris, computed over the final 1,000 episodes for all
  runs and the best single runs.}
\label{tab:resSZ}
\begin{center}
\begin{tabular}{l|c|c} 
\bf{Network} & \bf{Final average score} & \bf{Final best score }\\ 
\hline
\hline
\multicolumn{3}{c}{Shallow networks} \\
\hline
SiLU     & $214 \pm 74$ & $253 \pm 83$  \\
ReLU     & $191 \pm 58$ & $227 \pm 76$  \\
dSiLU   & $\mathbf{263 \pm 80}$ & $\mathbf{320 \pm 87}$ \\
Sigmoid & $232 \pm 75$ & $293 \pm 73$ \\
\hline
\multicolumn{3}{c}{Deep networks} \\
\hline
SiLU-SiLU    & $217 \pm 53$ & $219 \pm 54$  \\
ReLU-ReLU    & $215 \pm 54$ & $217 \pm 52$  \\
SiLU-dSiLU   & $\mathbf{229 \pm 55}$ & $\mathbf{235 \pm 54}$ \\
\hline
\end{tabular} 
\end{center}
\end{table}

Figure~\ref{fig:res_SZtet} shows the average learning curves as well
as learning curves for the individual runs for the shallow networks,
Figure~\ref{fig:res_deepSZtet} shows the average learning curves for
the deep networks, and the final results are summarized in
Table~\ref{tab:resSZ}. The results show significant differences ($p <
0.0001$) in final average score between all four shallow agents. The
networks with bounded hidden units (dSiLU and sigmoid) outperformed
the networks with unbounded units (SiLU and ReLU), the SiLU network
outperformed the ReLU network, and the dSiLU network outperformed the
sigmoid network. The final average score (best score) of 263 (320)
points achieved by the dSiLU network agent is a new state-of-the-art
score, improving the previous best performance by $43$ (25) points or
$20\,\%$ (8\,\%). In the deep learning setting, the SiLU-dSiLU network
significantly ($p < 0.0001$) outperformed the other two networks and
the average final score of 229 points is better than the previous
state-of-the-art of 220 points. There was no significant difference
($p = 0.32$) between the final performance of the SiLU-SiLU network
and the ReLU-ReLU network.

\subsection[10x10 Tetris]{10$\times$10 Tetris}
The result achieved by the dSiLU network agent in stochastic SZ-Tetris
is impressive, but we cannot compare the result with the methods that
have achieved the highest performance levels in standard Tetris
because those methods have not been applied to stochastic SZ-Tetris.
Furthermore, it is not feasible to apply our method to Tetris with a
standard board height of $20$, because of the prohibitively long
learning time. The current state-of-the-art for a single run of an
algorithm, achieved by the CBMPI
algorithm~\citep{Gabillon13,Scherrer15}, is a mean score of 51 million
cleared lines. However, for the best methods applied to Tetris, there
are reported results for a smaller, 10$\times$10, Tetris board,
and in this case, the learning time for our method is long, but not
prohibitively so.
\begin{figure}[!htb]
   \begin{center}
     \includegraphics[width=0.5\textwidth]{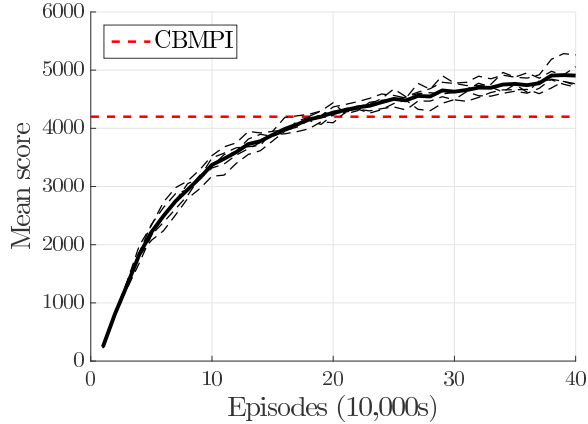}
  \end{center}
  \caption{Learning curves for a dSiLU network agent with 250 hidden
    nodes in 10$\times$10 Tetris. The figure shows the average score
    over five separate runs (tick solid lines) and the scores of
    individual runs (thin dashed lines). The red dashed line show the
    previous best average score of 4,200 points achieved by the CBMPI
    algorithm.}
  \label{fig:res_small_tet} 
\end{figure}

10$\times$10 Tetris is played with the standard seven tetrominoes and the
numbers of actions are 9 for the block-shaped tetromino, 17 for the S-,
Z-, and stick-shaped tetrominoes, and 34 for the J-, L- and T-shaped
tetrominoes. In each time step, the agent gets a score equal to the
number of completed rows, with a maximum of +4 points that can only be
achieved by the stick-shaped tetromino.

We trained a shallow neural network agent with dSiLU units in the hidden
layer. To handle the more complex learning task, we increased the
number of hidden units to 250 and the number of episodes to
400,000. We repeated the experiment for five separate runs. We used
the same 20 state features as in the SZ-Tetris experiment, but the
length of the binary state vector was reduced to 260 due to the
smaller board size. The reward function was changed as follows for the
same reason:
\begin{equation}
r(\mathbf{s}) = e^{-(\textrm{number of holes in }\mathbf{s})/(33/2)}.
\label{eq:tetris_r}
\end{equation}
We used the same values of the meta-parameters as in the stochastic
SZ-Tetris experiment.

Figure~\ref{fig:res_small_tet} shows The average learning curve in
10$\times$10 Tetris, as well as learning curves for the five separate
runs. The dSiLU network agent reached an average score of 4,900 points
over the final 10,000 episodes and the five separate runs, which is a
new state-of-the-art in 10$\times$10 Tetris. The previous best average
scores are 4,200 points achieved by the CBMPI algorithm, 3,400 points
achieved by the DPI algorithm, and 3,000 points achieved by the CE
method~\citep{Gabillon13}. The best individual run achieved a final
mean score of 5,300 points, which is also a new state-of-the-art,
improving on the score of 5,000 points achieved by the CBMPI
algorithm.

It is particularly impressive that the dSiLU network agent achieved
its result using features similar to the original Bertsekas
features. Using only the Bertsekas features, the CBMPI algorithm, the
DPI algorithm, and the CE method could only achieve average scores of
about 500 points~\citep{Gabillon13}. The CE method has achieved its
best score by combining the Bertsekas features, the Dellacherie
features~\citep{Fahey03}, and three original
features~\citep{Thiery09}. The CBMPI algorithm achieved its best score
using the same features as the CE method, except for using five
original RBF height features instead of the Bertsekas features.

\subsection{Atari 2600 games}
To further evaluate the use of value-based on-policy reinforcement
learning with eligibility traces and softmax action selection in
high-dimensional state space domains, as well as the use of SiLU and
dSiLU units, we applied Sarsa($\lambda$) with a deep convolution
neural network function approximator in the Atari 2600 domain using
the Arcade Learning Environment~\citep{Bellemare13}. Based on the
results for the deep networks in SZ-Tetris, we used SiLU-dSiLU
networks with SiLU units in the convolutional layers and dSiLU units in
the fully-connected layer. To limit the number of games and prevent a
biased selection of the games, we selected the 12 games played by
DQN~\citep{Mnih15} that started with the letters 'A' and 'B': Alien,
Amidar, Assault, Asterix, Asteroids, Atlantis, Bank Heist, Battle
Zone, Beam Rider, Bowling, Boxing, and Breakout.

We used a similar experimental setup as~\citet{Mnih15}. We
pre-processed the raw 210$\times$160 Atari 2600 RGB frames by extracting
the luminance channel, taking the maximum pixel values over
consecutive frames to prevent flickering, and then downsampling the
grayscale images to 105$\times$80. For computational reasons, we used a
smaller network architecture. Instead of three convolutional layers,
we used two with half the number of filters, each followed by a
max-pooling layer. The input to the network was a 105$\times$80$\times$2 image
consisting of the current and the fourth previous pre-processed
frame. As we used frame skipping where actions were selected every
fourth frame and repeated for the next four frames, we only needed to
apply pre-processing to every fourth frame. The first convolutional
layer had 16 filters of size 8$\times$8 with a stride of 4. The second
convolutional layer had 32 filters of size 4$\times$4 with a stride of
2. The max-pooling layers had pooling windows of size 3$\times$3 with a
stride of 2. The convolutional layers were followed by a
fully-connected hidden layer with 512 dSiLU units and a
fully-connected linear output layer with 4 to 18 output (or
action-value) units, depending on the number of valid actions in the
considered game.
\begin{figure}[!ht]
   \begin{center}
    \subfigure
    {
      \includegraphics[width=0.31\textwidth]{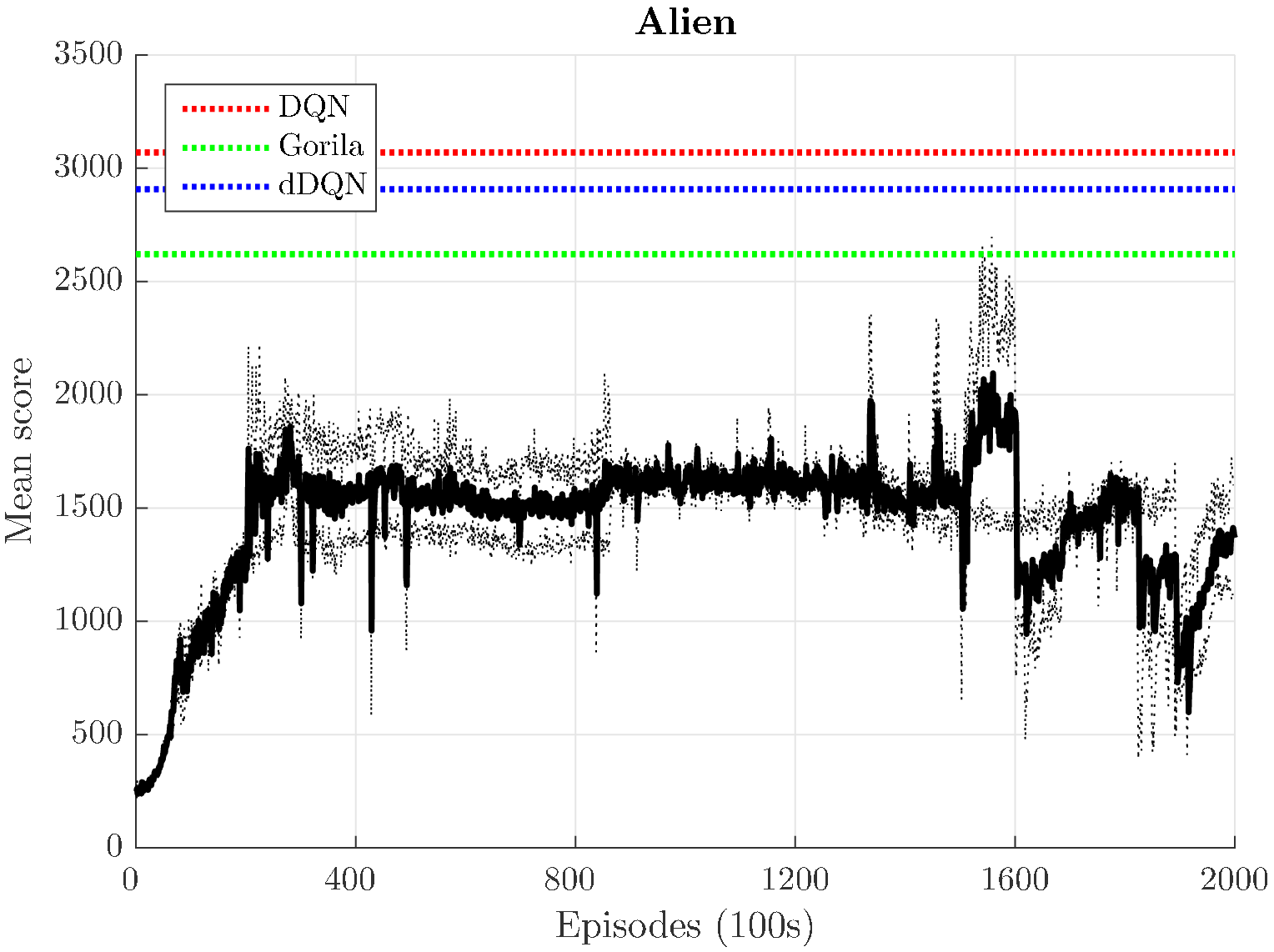} 
    } 
    \subfigure
    {
      \includegraphics[width=0.31\textwidth]{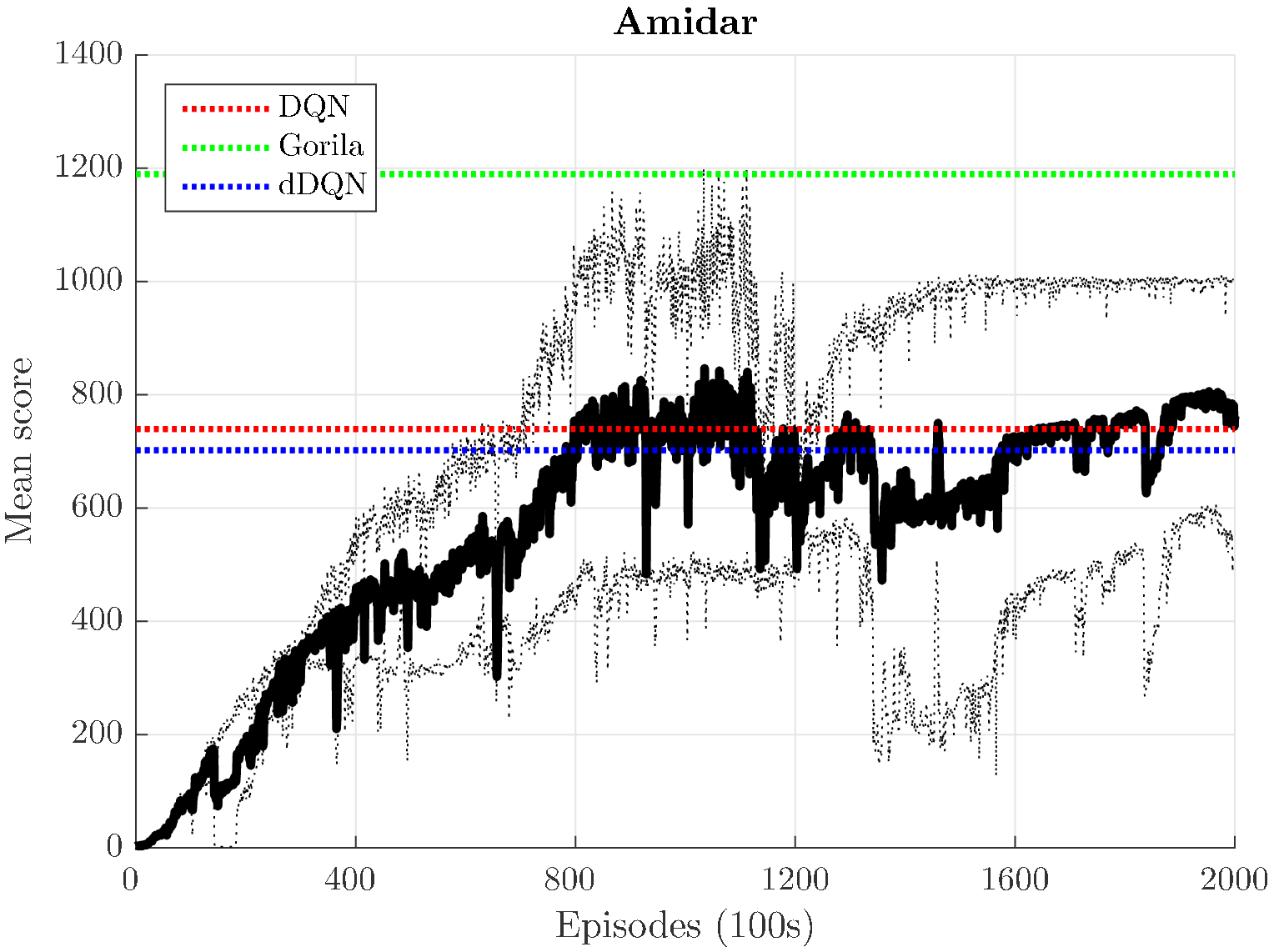} 
    } 
    \subfigure
    {
      \includegraphics[width=0.31\textwidth]{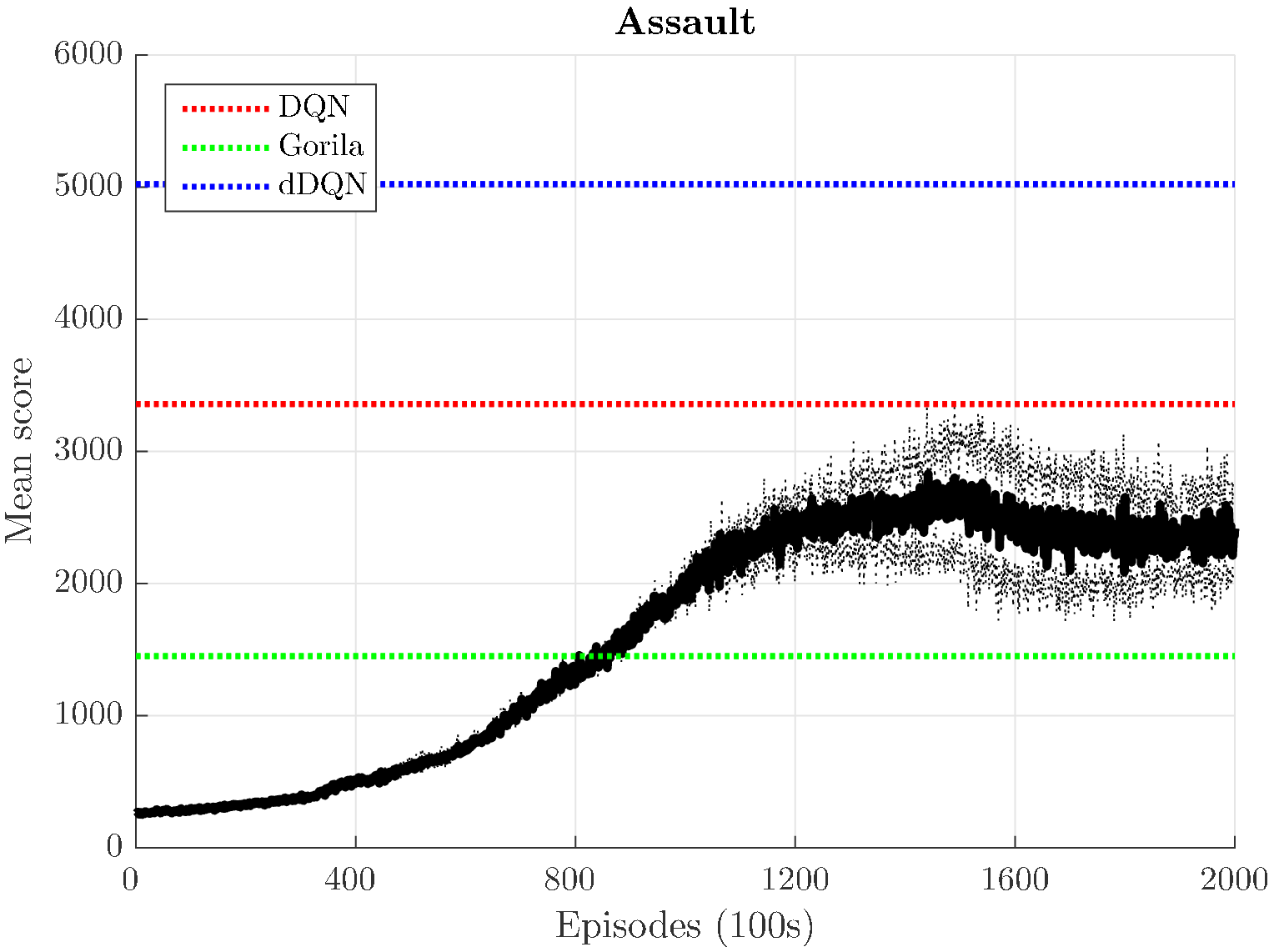} 
    } 
    \subfigure
    {
      \includegraphics[width=0.31\textwidth]{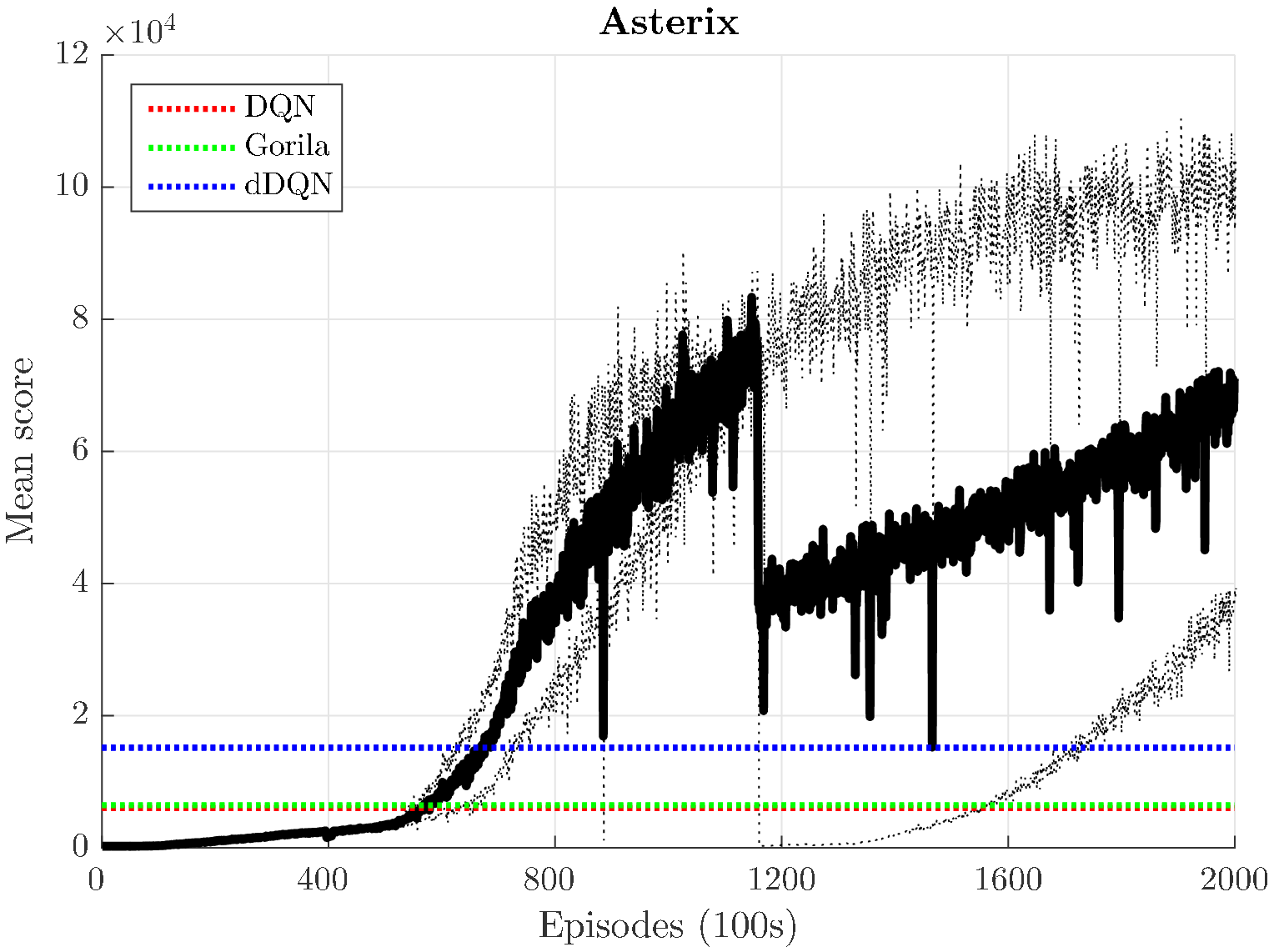} 
    }
    \subfigure
    {
      \includegraphics[width=0.31\textwidth]{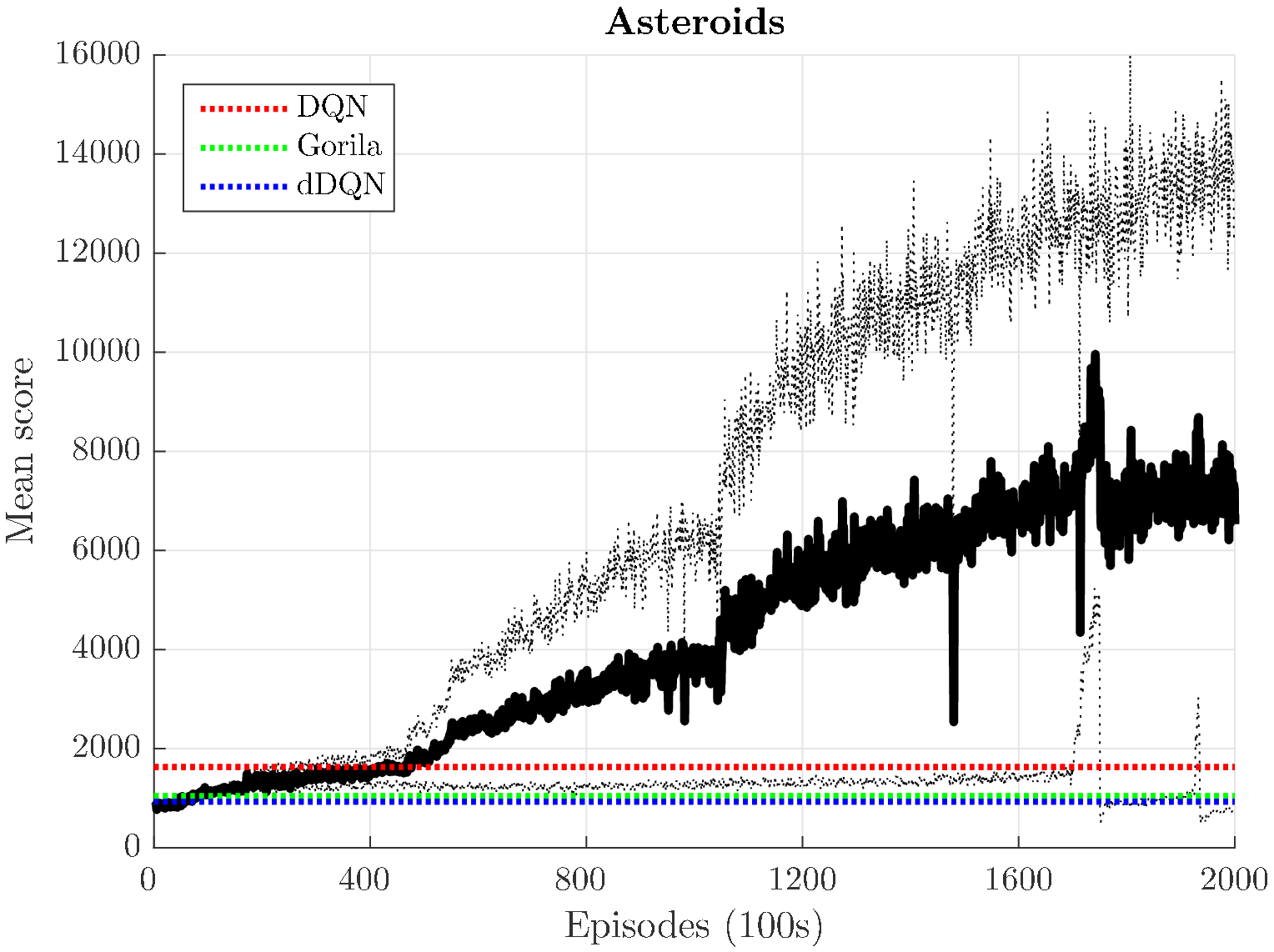} 
    } 
    \subfigure
    {
      \includegraphics[width=0.31\textwidth]{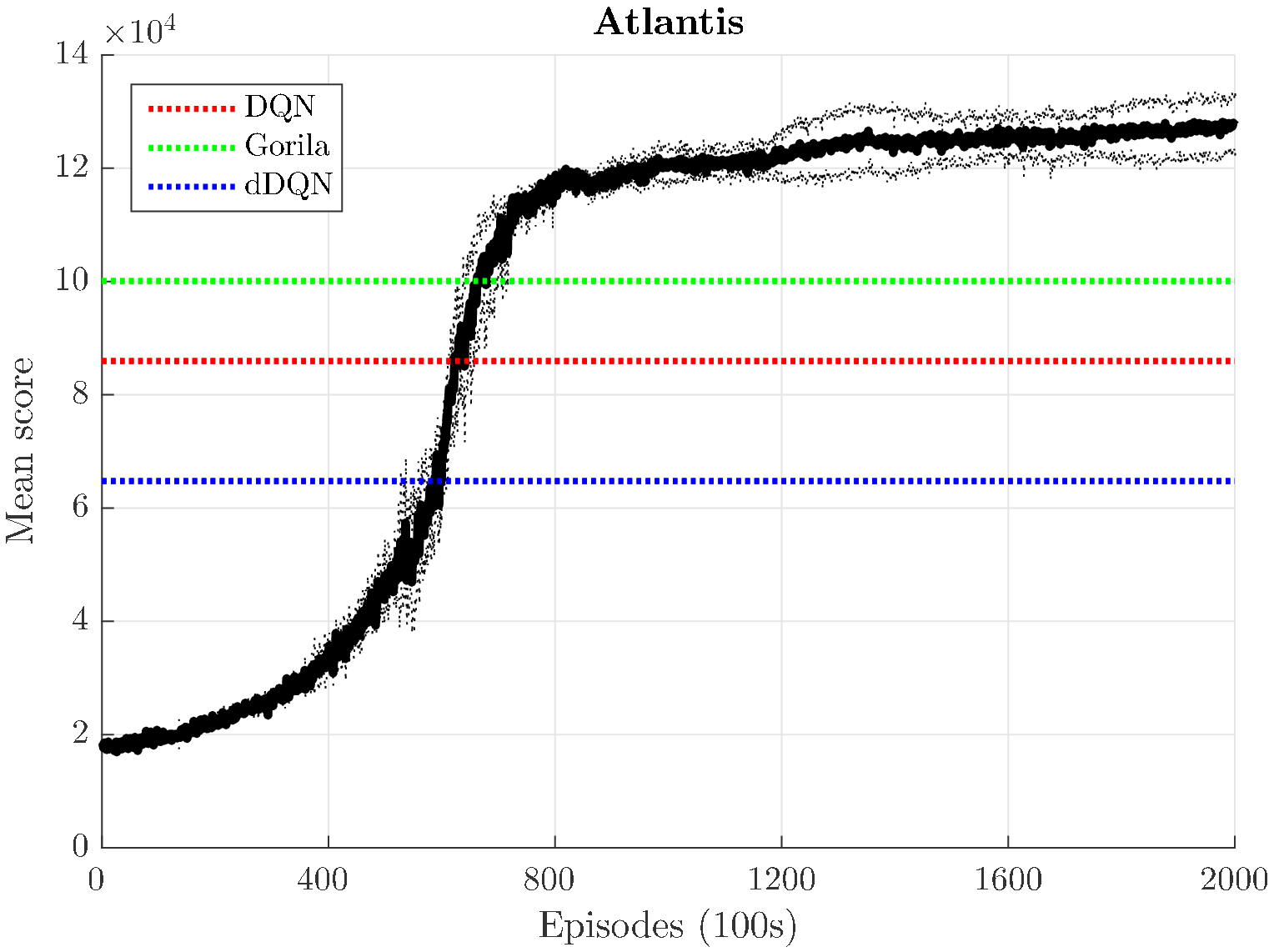} 
    } 
    \subfigure
    {
      \includegraphics[width=0.31\textwidth]{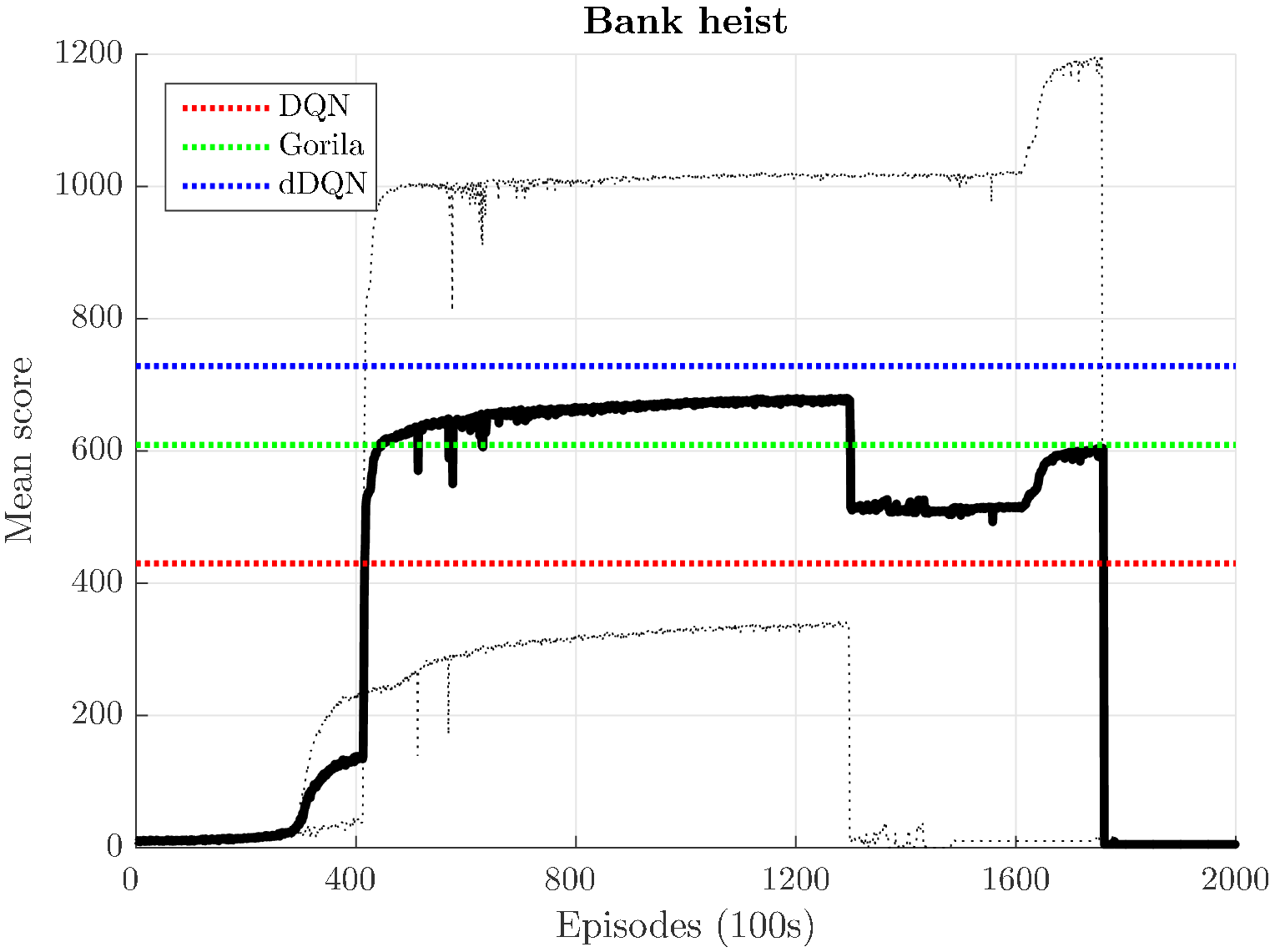} 
    }
    \subfigure
    {
      \includegraphics[width=0.31\textwidth]{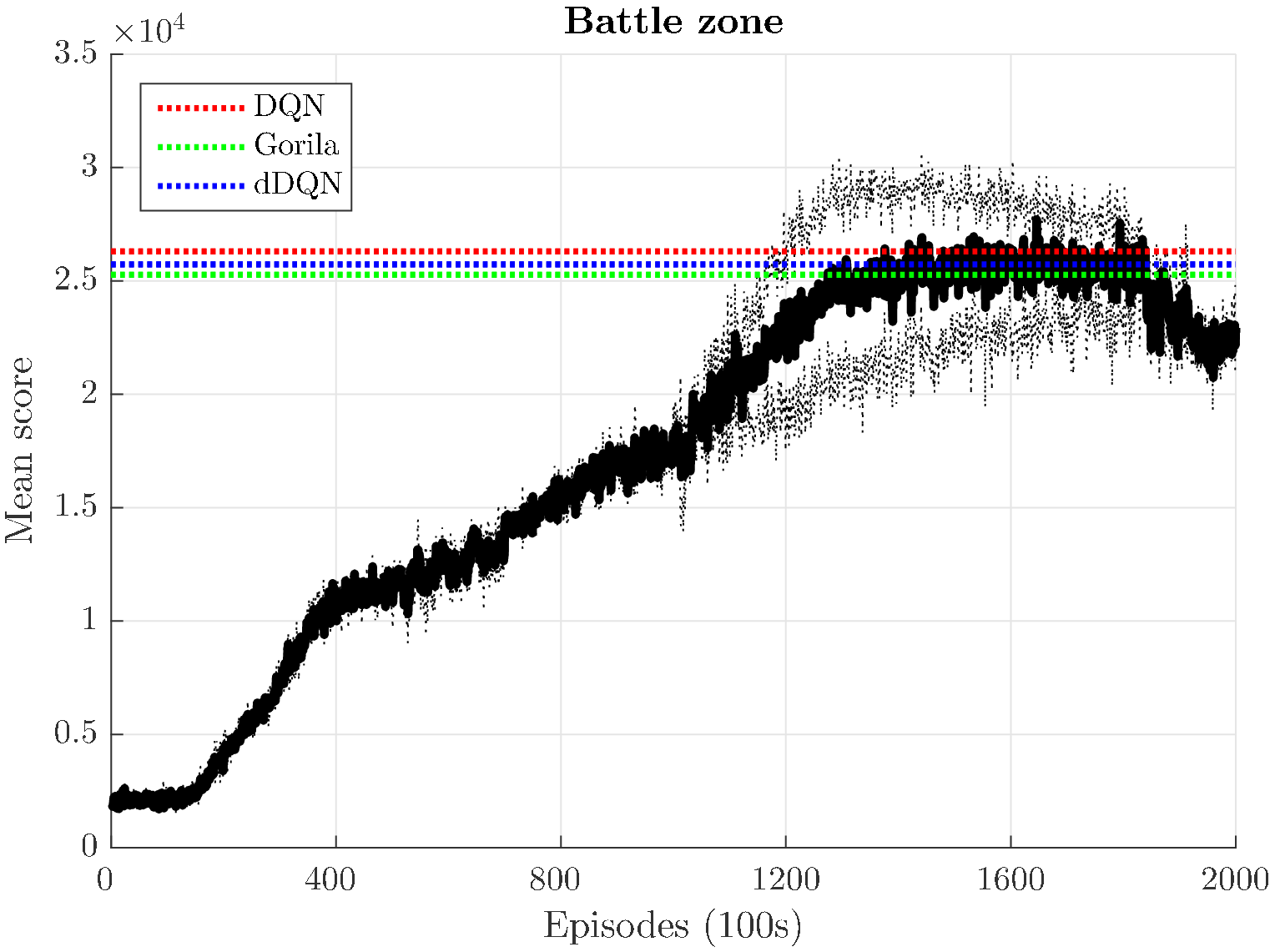} 
    } 
    \subfigure
    {
      \includegraphics[width=0.31\textwidth]{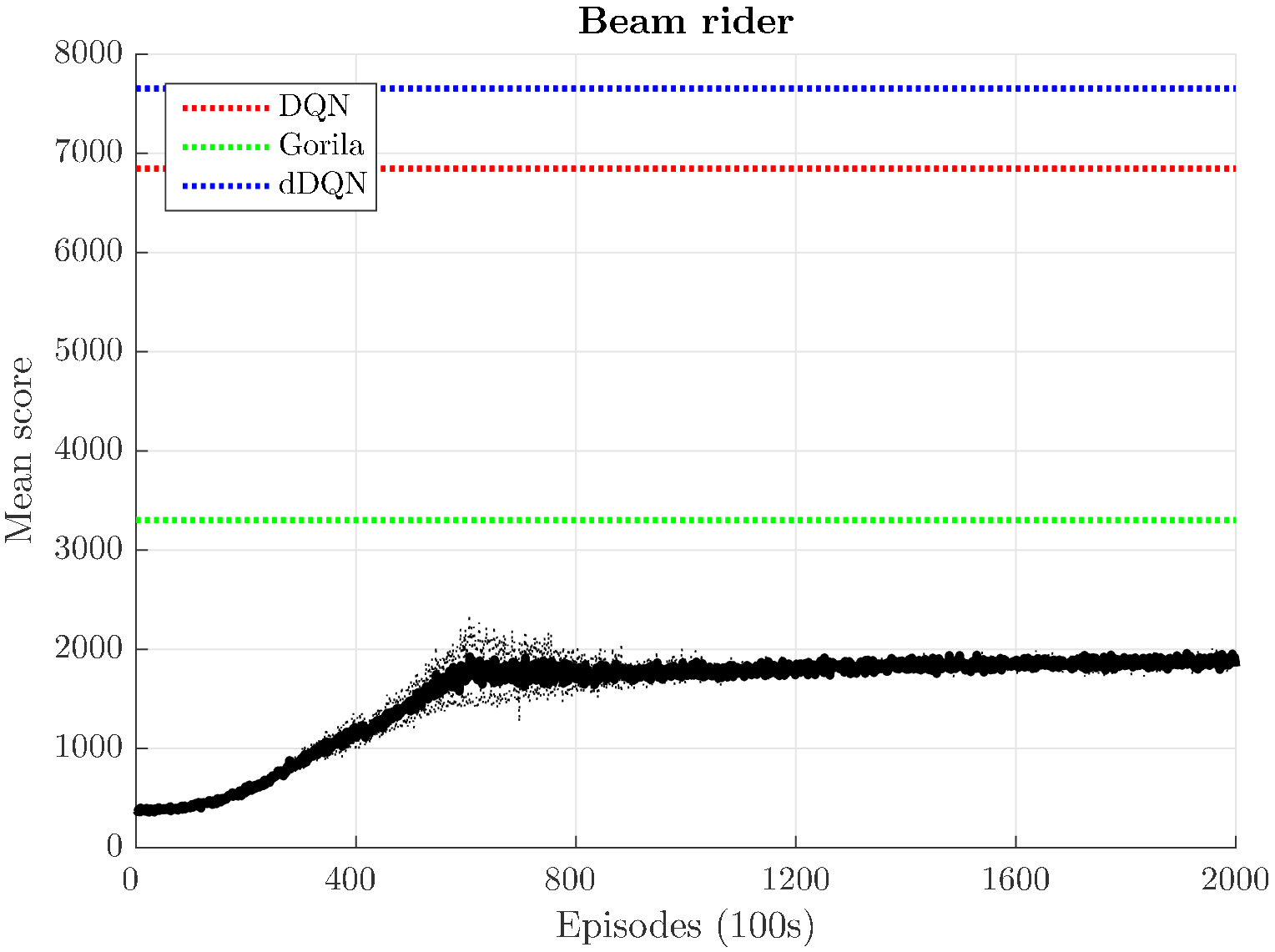} 
    } 
    \subfigure
    {
      \includegraphics[width=0.31\textwidth]{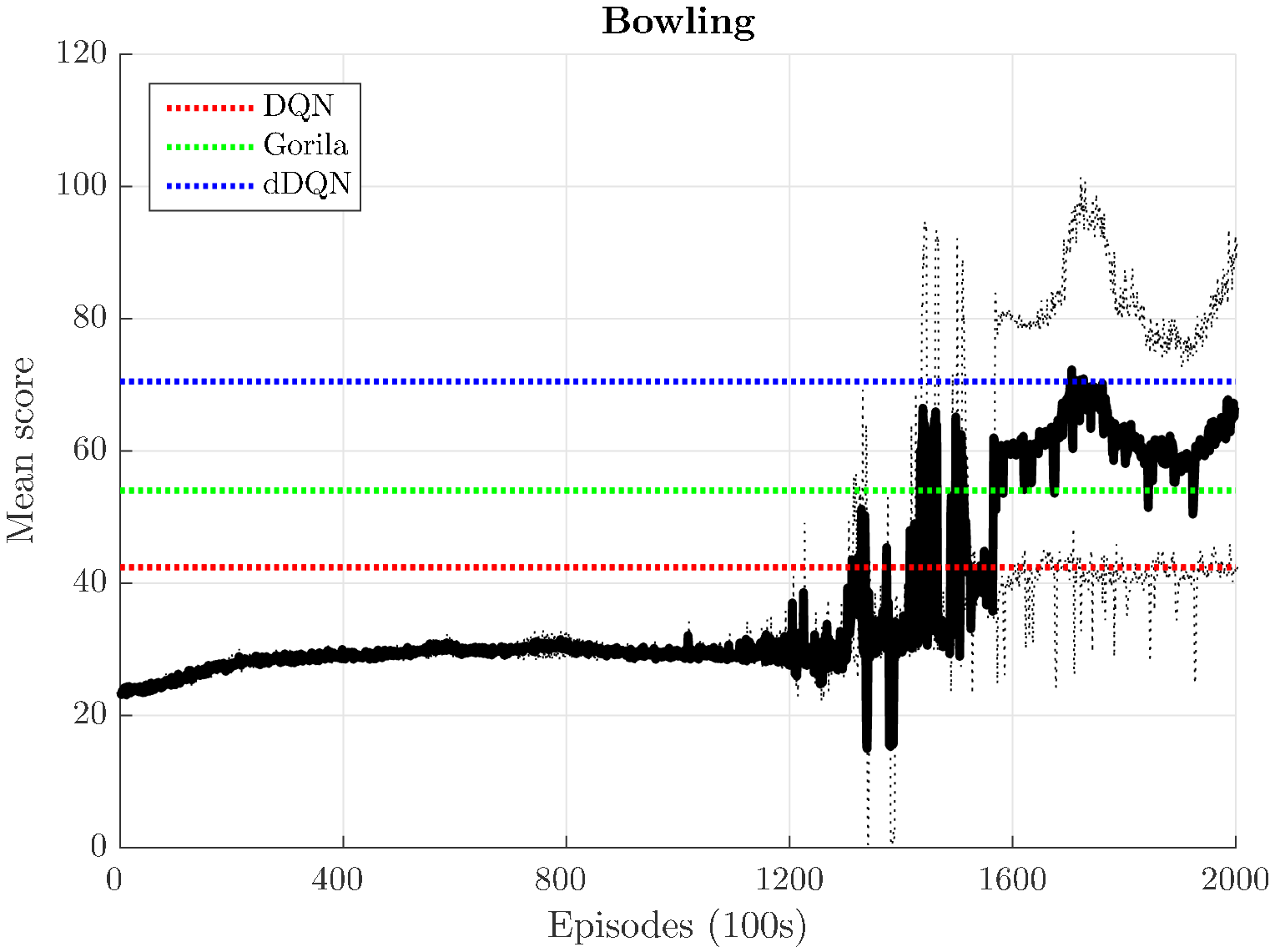} 
    }
    \subfigure
    {
      \includegraphics[width=0.31\textwidth]{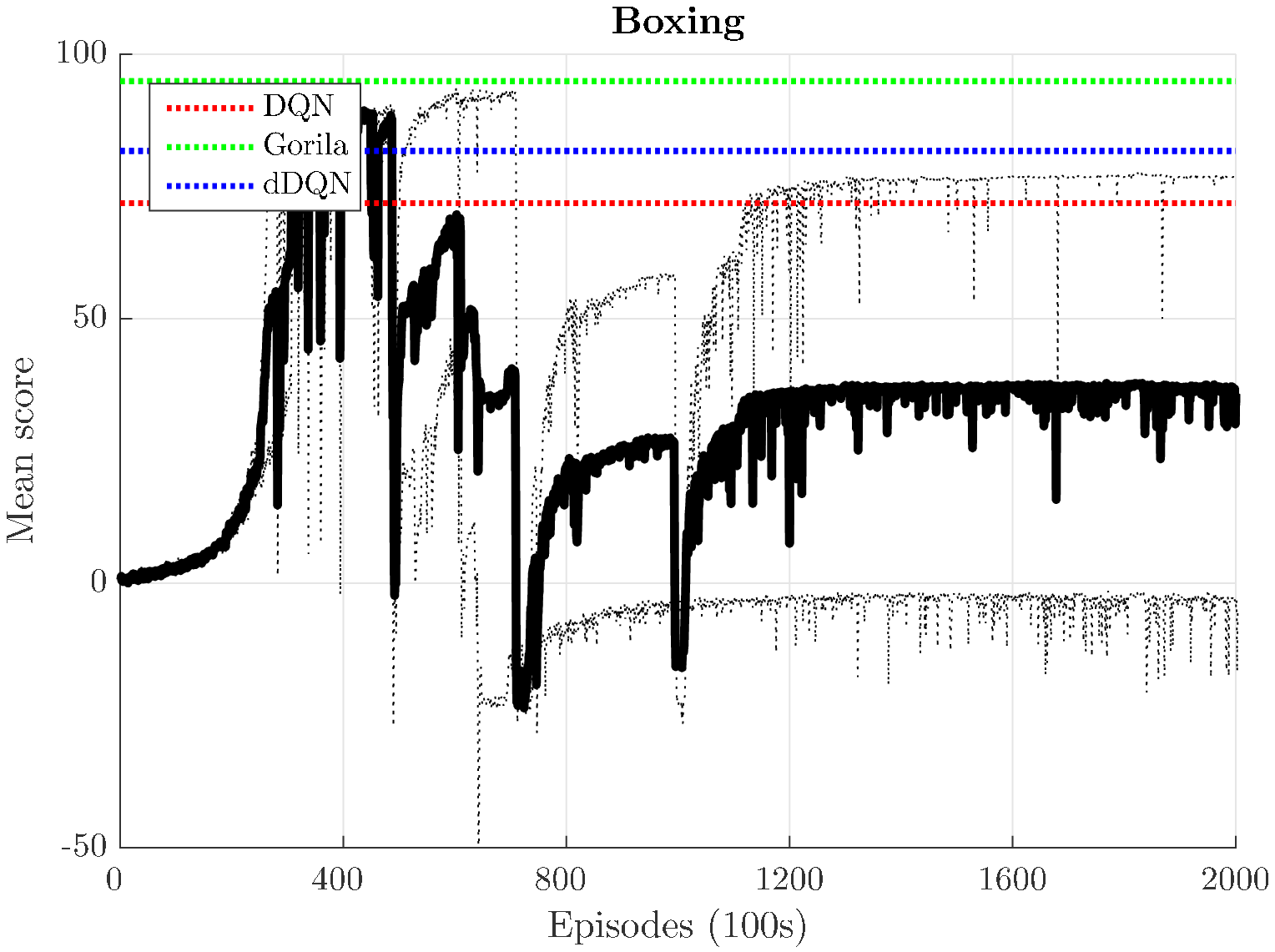} 
    } 
    \subfigure
    {
      \includegraphics[width=0.31\textwidth]{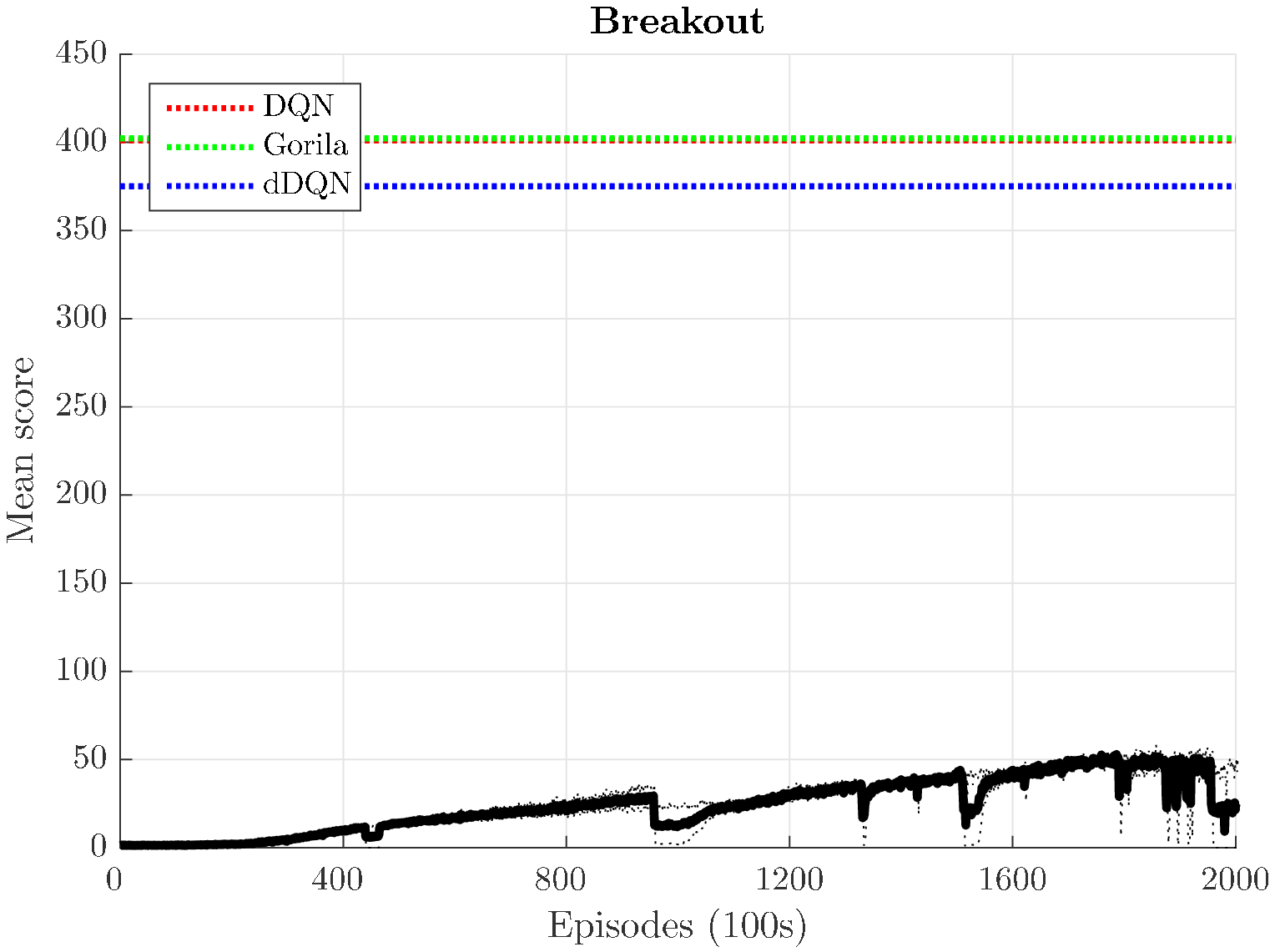} 
    } 
  \end{center}
  \caption{Average learning curves (solid lines) over two separate
    runs (dashed lines) for the SiLU-dSiLU agents in the 12 Atari
    games. The dotted lines show the reported results for DQN (red),
    the Gorila implementation of DQN (green), and double DQN (blue).}
  \label{fig:atari_res} 
\end{figure}
We selected meta-parameters by a preliminary search in the Alien,
Amidar and Assault games and used the same values for all 12 games:
$\alpha$: 0.001, $\gamma$: $0.99$, $\lambda$: $0.8$, $\tau_0$: $0.5$,
and $\tau_k$: $0.0005$. As in~\citet{Mnih15}, we clipped the rewards
to be between $-1$ and $+1$, but we did not clip the values of the
TD-errors.

In each of the 12 Atari games, we trained a SiLU-dSiLU agent for
200,000 episodes and the experiments were repeated for two separate
runs. An episode started with up to $30$ 'do nothing' actions
(\emph{no-op condition}) and it was played until the end of the game
or for a maximum of 18,000 frames (i.e., 5 minutes).
Figure~\ref{fig:atari_res} shows the average learning curves, as well
as the learning curves for the two separate runs, in the 12 Atari 2600
games. Table~\ref{tab:atari_res} summarizes the results as the
\emph{final mean scores} computed over the final 100 episodes for the
two separate runs, and the \emph{best mean scores} computed as average
scores of the highest mean scores (over every 100 episodes) achieved
in each of the two runs. The table also shows the reported best mean
scores for single runs of DQN computed over 30 episodes, average
scores over five separate runs of the Gorila implementation of
DQN~\citep{Nair15} computed over 30 episodes, and single runs of
double DQN~\citep{Hasselt15} computed over 100 episodes. The last two
rows of the table shows summary statistics over the 12 games, which
were obtained by computing the mean and the median of the DQN
normalized scores:
\begin{displaymath}
\mathrm{Score_{DQN\_normalized}} = \frac{\mathrm{Score_{agent}} -
  \mathrm{Score_{random}} }{\mathrm{Score_{DQN}} -
  \mathrm{Score_{random}}}
\end{displaymath}
Here, $\mathrm{Score_{random}}$ is the score achieved by a random
agent in~\citet{Mnih15}.

\begin{table*}[!htb]
\caption{The final and best mean scores achieved by the SiLU-dSiLU
  agents in 12 Atari 2600 games, and the reported best mean scores
  achieved by DQN, the Gorila implementation of DQN, and double DQN in
  the no-op condition with 5 minutes of evaluation time.}
\label{tab:atari_res}
\begin{center}
\begin{tabular}{l|c|c|c||c|c} 
 {}        &     {}   &    {}       &     {}          & \multicolumn{2}{c}{\bf{SiLU-dSiLU}} \\ 
\bf{Game}  & \bf{DQN} & \bf{Gorila} & \bf{double DQN} & \bf{Final} & \bf{Best}\\ 
\hline
\hline
Alien       & \bf{3,069}  & 2,621   & 2,907  & 1,370   &   2,246   \\
Amidar      & 740   & \bf{1,190}   & 702   & 762    &   904    \\
Assault     & 3,359  & 1,450   & \bf{5,023}  & 2,415   &   2,944   \\
Asterix     & 6,012  & 6,433   & 15,150 & 70,942  &   \bf{100,322} \\
Asteroids   & 1,629  & 1,048   & 931   & 6,537   &   \bf{10,614}  \\
Atlantis    & 85,950 & 100,069 & 64,758 & 127,651 &   \bf{128,983} \\
Bank Heist  & 430   & 609    & 728   & 5      &   \bf{770}    \\
Battle Zone & 26,300 & 25,267  & 25,730 & 22,930  &   \bf{29,115}  \\
Beam Rider  & 6,846  & 3,303   & \bf{7,654}  & 1,829   &   2,176        \\
Bowling     & 42    & 54     & 71    & 67     &   \bf{75}     \\
Boxing      & 72    & \bf{95}     & 82    & 36     &   92      \\
Breakout    & 401   & \bf{402}    & 375   & 25     &   55     \\
\hline
{\bf Mean} (DQN Normalized)   & $100\,\%$  & $102\,\%$    & $127\,\%$   & $218\,\%$  & $\mathbf{332\,\%}$  \\
{\bf Median} (DQN Normalized) & $100\,\%$ & $104\,\%$    & $105\,\%$   & $78\,\%$  &   $\mathbf{125\,\%}$    \\
\end{tabular} 
\end{center}
\end{table*}
The results clearly show that our SiLU-dSiLU agent outperformed the
other agents, improving the mean (median) DQN normalized best mean
score from $127\,\%$ ($105\,\%$)  achieved by double DQN to $332\,\%$
($125\,\%$). The SiLU-dSiLU agents achieved the highest best mean score
in 6 out of the 12 games and only performed much worse than the other
3 agents in one game, Breakout, where the learning never took off
during the 200,000 episodes of training (see
Figure~\ref{fig:atari_res}). The performance was especially impressive
in the Asterix (score of 100,322) and Asteroids (score of 10,614)
games, which improved the best mean performance achieved by the
second-best agent by $562\,\%$ and $552\,\%$, respectively.

\section{Analysis}
\subsection{Value estimation}
First, we investigate the ability of TD($\lambda$) and
Sarsa($\lambda$) to accurately estimate discounted returns: 
\begin{displaymath}
R_t =\sum_{k=0}^{T-t} \gamma^kr_{t+k}.
\end{displaymath}
Here $T$ is the length of an
episode. The reason for doing this is that \citet{Hasselt15} showed
that the double DQN algorithm improved the performance of DQN in Atari
2600 games by reducing the overestimation of the action values. It is
known~\citep{Thrun93,Hasselt10} that Q-learning based algorithms, such
as DQN, can overestimate action values due to the max operator, which
is used in the computation of the learning targets. TD($\lambda$) and
Sarsa($\lambda$) do not use the max operator to compute the learning
targets and they should therefore not suffer from this problem.

\begin{figure}[!htb]
   \begin{center}
     \subfigure
         {     
           \includegraphics[width=0.47\textwidth]{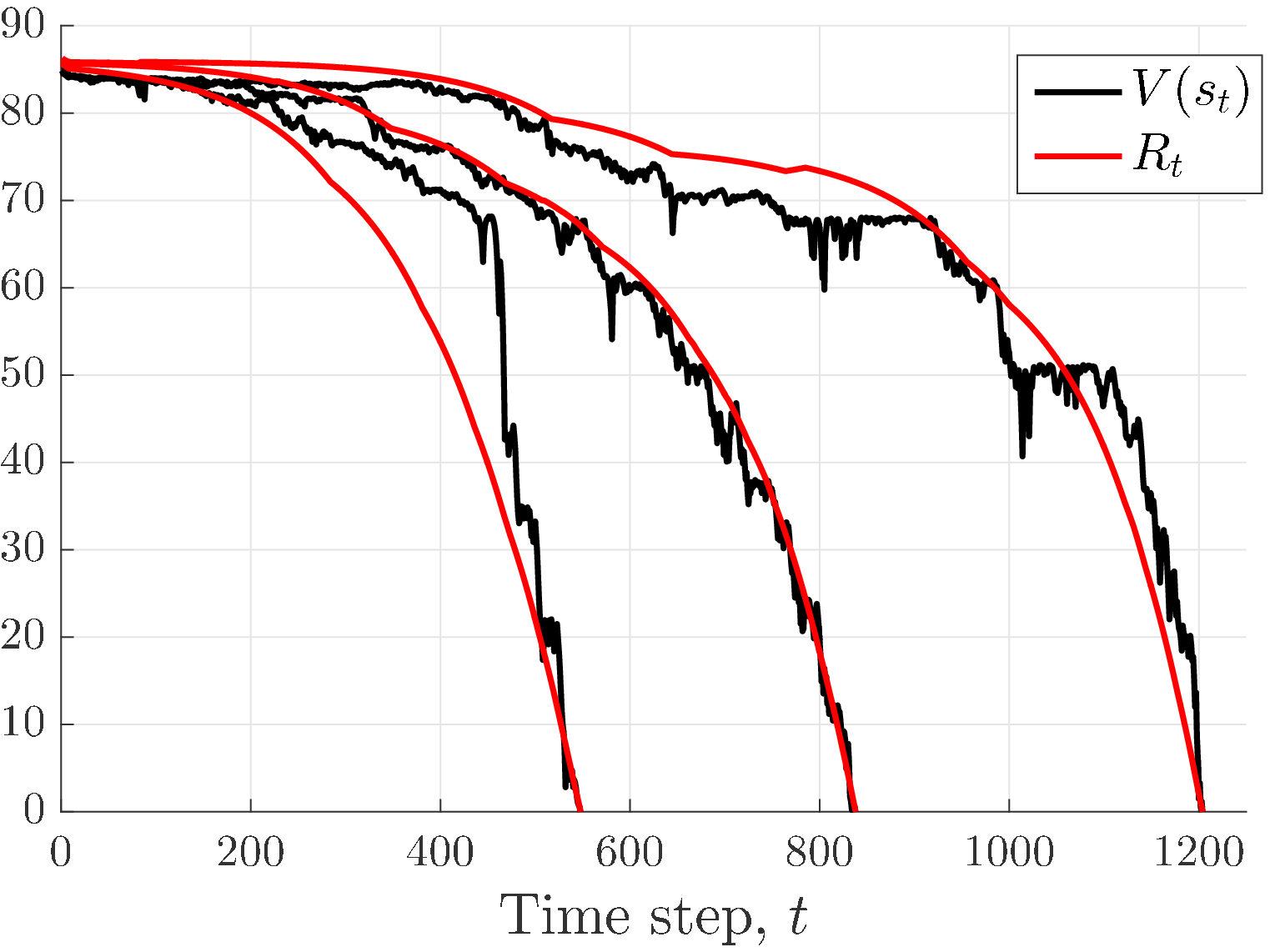}
         }
     \subfigure
         {     
           \includegraphics[width=0.47\textwidth]{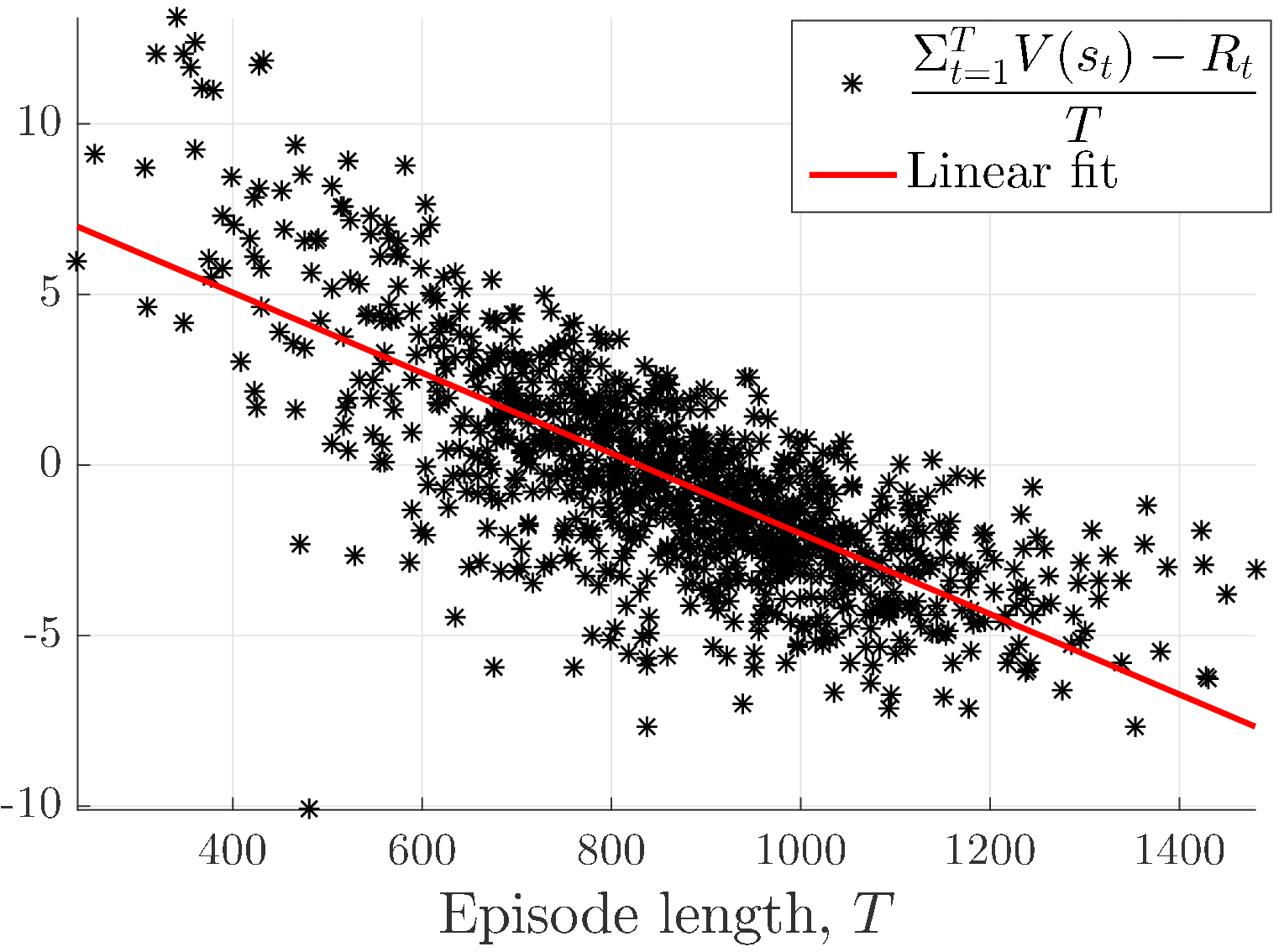}
         }
  \end{center}
  \caption{The left panel shows learned $V(s_t)$-values and
    $R_t$-values, for examples of short, medium-long, and long
    episodes in SZ-Tetris. The right panel shows the normalized sum of
    differences between $V(s_t)$ and $R_t$ for 1,000 episodes and the
    best linear fit of the data ($-0.012T + 9.8$).}
  \label{fig:Vs_v_Rt_SZtet} 
\end{figure}
Figure~\ref{fig:Vs_v_Rt_SZtet} shows that for episodes of average (or
expected) length the best dSiLU network agent in SZ-Tetris learned
good estimates of the discounted returns, both along the episodes
(left panel) and as measured by the normalized sum of differences
between $V(s_t)$ and $R_t$ (right panel):
\begin{displaymath}
\frac{1}{T}\sum\nolimits_{t=1}^T \left(V(s_t) - R_t\right).
\end{displaymath}
The linear fit of the normalized sum of differences data for 1,000
episodes gives a small underestimation (-0.43) for an episode of
average length (866 time steps). The $V(s_t)$-values overestimated the
discounted returns for short episodes and underestimated the discounted
returns for long episodes (especially in the middle part of the
episodes), which is accurate since the episodes ended earlier and
later, respectively, than were expected.

\begin{figure}[!htb]
   \begin{center}
     \subfigure
         {     
           \includegraphics[width=0.95\columnwidth]{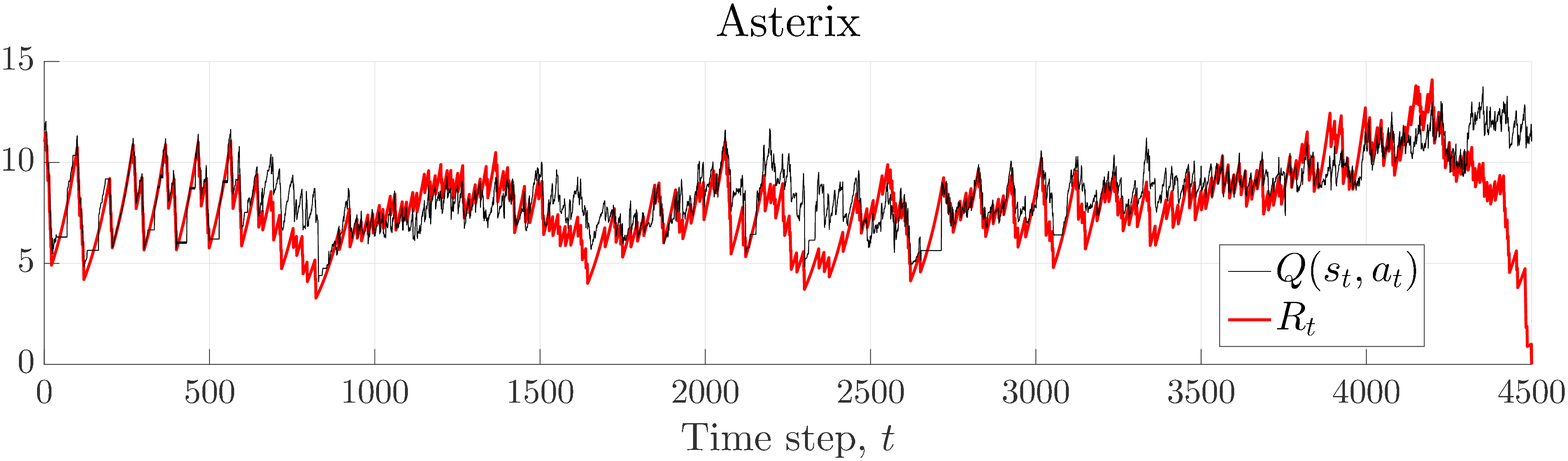}
         }
     \subfigure
         {     
           \includegraphics[width=0.95\columnwidth]{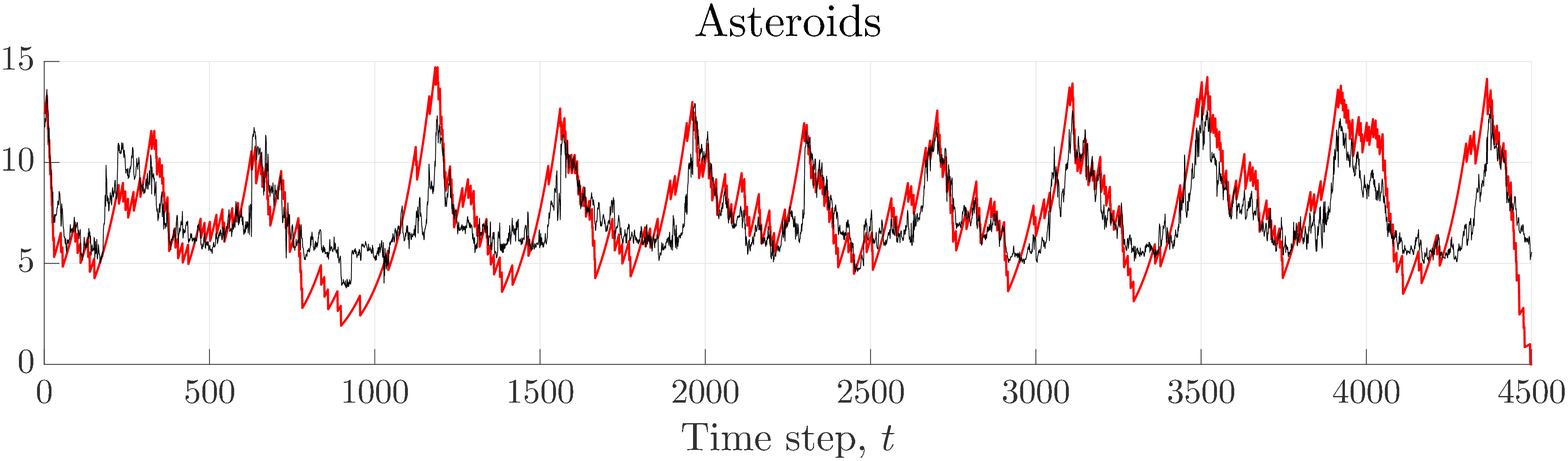}
         }
  \end{center}
  \caption{Learned action values, $Q(s_t, a_t)$, and discounted
    returns, $R_t$, for the best SiLU-dSiLU agents in Asterix and
    Asteroids.}
  \label{fig:Qsa_Rt_atari}
\end{figure}
Figure~\ref{fig:Qsa_Rt_atari} shows typical examples of learned action
values and discounted returns along episodes where the best SiLU-dSiLU
agents in Asterix (score of 108,500) and Asteroids (score of 22,500)
successfully played for the full 18,000 frames (i.e., 4,500 time
steps since the agents acted every fourth frame). In both games, with
the exception of a few smaller parts, the learned action values
matched the discounted returns very well along the whole episodes. The
normalized sums of differences (absolute differences) were $0.59$
($1.05$) in the Asterix episode and $-0.23$ ($1.28$) in the Asteroids
episode. In both games, the agents overestimated action values at the
end of the episodes. However, this is an artifact of that an episode
ended after a maximum of 4,500 time steps, which the agents could not
predict. Videos of the corresponding learned behaviors in Asterix and
Asteroids can be found at \url{http://www.cns.atr.jp/~elfwing/videos/asterix_deep_SiL.mov} and \\\url{http://www.cns.atr.jp/~elfwing/videos/asteroids_deep_SiL.mov}.
\begin{table}[!htb]
\caption{Mean scores and average numbers of exploratory actions for
  softmax action selection and $\varepsilon$-greedy action selection
  with $\varepsilon$ set to 0, 0.001, 0.01, and 0.05.}
\label{tab:egreedy_softmax}
\begin{center}
\begin{tabular}{l|l|c|c} 
     &   &   \bf{Mean} & \bf{Exploratory} \\ 
\bf{Game} & \bf{Selection} & \bf{Score} & \bf{actions} \\ 
\hline
\hline 
\multirow{4}{*}{SZ-Tetris} 
 & $\tau     = 0.0098$            & 326 & 28.7 \\ 
 & $\epsilon = 0$                 & 332 & 0         \\ 
 & $\varepsilon = 0.001$             & 260 & 0.59   \\ 
 & $\varepsilon = 0.01$              & 71  & 2.0    \\ 
 & $\varepsilon = 0.05$              & 14  & 3.2   \\
\hline
\hline
\multirow{4}{*}{Asterix} 
 & $\tau     = 0.00495$            & 104,299  &  47.6   \\
 & $\epsilon = 0$                  & 102,890 & 0      \\  
 & $\varepsilon = 0.001$              & 98,264 & 3.6    \\ 
 & $\varepsilon = 0.01$               & 66,113 & 30.0  \\ 
 & $\varepsilon = 0.05$               & 7,152  & 56.8  \\
\hline
\hline
\multirow{4}{*}{Asteroids} 
 & $\tau     = 0.00495$              & 15,833 & 31.3   \\
 & $\epsilon = 0$                    & 15,091 &  0        \\ 
 & $\varepsilon = 0.001$                & 11,105 & 2.1    \\ 
 & $\varepsilon = 0.01$                 & 3,536  & 11.7     \\ 
 & $\varepsilon = 0.05$                 & 1,521  & 47.3     \\
\end{tabular} 
\end{center}
\end{table}
\subsection{Action selection}
Second, we investigate the importance of softmax action selection in
the games where our proposed agents performed particularly
well. Almost all deep reinforcement learning algorithms that have been
used in the Atari 2600 domain have used $\varepsilon$-greedy action
selection (one exception is the asynchronous advantage actor-critic
method, A3C, which used softmax output units for the
actor~\citep{Mnih2016}). One drawback of $\varepsilon$-greedy
selection is that it selects all actions with equal probability when
exploring, which can lead to poor learning outcomes in tasks where the
worst actions have very bad consequences. This is clearly the case in
both Tetris games and in the Asterix and Asteroids games. In each
state in Tetris, many, and often most, actions will create holes,
which are difficult (especially in SZ-Tetris) to remove. In the
Asterix game, random exploratory actions can kill Asterix if executed
when Cacofonix's deadly lyres are passing. In the Asteroids game, one
of the actions sends the spaceship into hyperspace and makes it
reappear in a random location, which has the risk of the spaceship
self-destructing or of destroying it by appearing on top of an asteroid.

We compared softmax action selection ($\tau$ set to the final values)
and $\varepsilon$-greedy action selection with $\varepsilon$ set to 0,
0.001, 0.01, and 0.05 for the best dSiLU network agent in SZ-Tetris and
the best SiLU-dSiLU agents in the Asterix and Asteroids games. The
results (see Table~\ref{tab:egreedy_softmax}) clearly show that
$\varepsilon$-greedy action selection with $\varepsilon$ set to 0.05,
as used for evaluation by DQN, is not suitable for these games. The
scores were only 4\,\% to 10\,\% of the scores for softmax
selection. The negative effects of random exploration were largest in
Asteroid and SZ-Tetris. Even when $\varepsilon$ was set as low as
0.001 and the agent performed only 2.1 exploratory actions per episode
in Asteroids and 0.59 in SZ-Tetris, the mean scores were reduced by
30\,\% and 20\,\% (26\,\% and 22\,\%), respectively, compared with
softmax selection ($\varepsilon = 0$).

\section{Conclusions}
In this study, we proposed SiLU and dSiLU as activation functions for
neural network function approximation in reinforcement learning. We
demonstrated in stochastic SZ-Tetris that SiLUs significantly
outperformed ReLUs, and that dSiLUs significantly outperformed sigmoid
units. The best agent, the dSiLU network agent, achieved new
state-of-the-art results in both stochastic SZ-Tetris and 10$\times$10
Tetris. In the Atari 2600 domain, a deep Sarsa($\lambda$) agent with
SiLUs in the convolutional layers and dSiLUs in the fully-connected
hidden layer outperformed DQN and double DQN, as measured by mean and
median DQN normalized scores.

An additional purpose of this study was to demonstrate that a more
traditional approach of using on-policy learning with eligibility
traces and softmax selection (i.e., basically a ``textbook'' version
of a reinforcement learning agent but with non-linear neural network
function approximators) can be competitive with the approach used by
DQN. This means that there is a lot of room for improvements, by,
e.g., using, as DQN, a separate target network, but also by using more
recent advances such as the dueling architecture~\citep{Wang2016} for
more accurate estimates of the action values and asynchronous learning
by multiple agents in parallel~\citep{Mnih2016}.

\section*{Acknowledgments}
This work was supported by the project commissioned by the New Energy
and Industrial Technology Development Organization (NEDO), JSPS
KAKENHI grant 16K12504, and Okinawa Institute of Science and
Technology Graduate University research support to KD.

\bibliography{learning}

\begin{thebibliography}{}

\bibitem[Bellemare et~al., 2013]{Bellemare13}
Bellemare, M.~G., Naddaf, Y., Veness, J., and Bowling, M. (2013).
\newblock The arcade learning environment: An evaluation platform for general
  agents.
\newblock {\em Journal of Artificial Intelligence Research}, 47:253--279.

\bibitem[Bertsekas and Ioffe, 1996]{Bertsekas96}
Bertsekas, D.~P. and Ioffe, S. (1996).
\newblock Temporal differences based policy iteration and applications in
  neuro-dynamic programming.
\newblock Technical Report LIDS-P-2349, MIT.

\bibitem[Burgiel, 1997]{Burgiel97}
Burgiel, H. (1997).
\newblock How to lose at {T}etris.
\newblock {\em Mathematical Gazette}, 81:194--200.

\bibitem[Elfwing et~al., 2015]{Elfwing15}
Elfwing, S., Uchibe, E., and Doya, K. (2015).
\newblock Expected energy-based restricted boltzmann machine for
  classification.
\newblock {\em Neural Networks}, 64(3):29--38.

\bibitem[Elfwing et~al., 2016]{Elfwing16}
Elfwing, S., Uchibe, E., and Doya, K. (2016).
\newblock From free energy to expected energy: Improving energy-based value
  function approximation in reinforcement learning.
\newblock {\em Neural Networks}, 84:17--27.

\bibitem[Fahey, 2003]{Fahey03}
Fahey, C. (2003).
\newblock Tetris {AI}, computer plays tetris.
\newblock \url{colinfahey.com/tetris/tetris.html} [Online; accessed
  22-February-2017].

\bibitem[Fau{\ss}er and Schwenker, 2013]{Fausser13}
Fau{\ss}er, S. and Schwenker, F. (2013).
\newblock Neural network ensembles in reinforcement learning.
\newblock {\em Neural Processing Letters}, pages 1--15.

\bibitem[Freund and Haussler, 1992]{Freund92}
Freund, Y. and Haussler, D. (1992).
\newblock Unsupervised learning of distributions on binary vectors using two
  layer networks.
\newblock In {\em Proceedings of Advances in Neural Information Processing
  Systems (NIPS1992)}.

\bibitem[Gabillon et~al., 2013]{Gabillon13}
Gabillon, V., Ghavamzadeh, M., and Scherrer, B. (2013).
\newblock Approximate dynamic programming finally performs well in the game of
  tetris.
\newblock In {\em Proceedings of Advances in Neural Information Processing
  Systems (NIPS2013)}, pages 1754--1762.

\bibitem[Hahnloser et~al., 2000]{Hahnloser00}
Hahnloser, R. H.~R., Sarpeshka, R., Mahowald, M.~A., Douglas, R.~J., and Seung,
  H.~S. (2000).
\newblock Digital selection and analogue amplification coexist in a
  cortex-inspired silicon circuit.
\newblock {\em Nature}, 405:947--951.

\bibitem[Hinton, 2002]{Hinton02}
Hinton, G.~E. (2002).
\newblock Training products of experts by minimizing contrastive divergence.
\newblock {\em Neural Computation}, 12(8):1771--1800.

\bibitem[Jaskowski et~al., 2015]{Jaskowski15}
Jaskowski, W., Szubert, M.~G., Liskowski, P., and Krawiec, K. (2015).
\newblock High-dimensional function approximation for knowledge-free
  reinforcement learning: a case study in {SZ}-{T}etris.
\newblock In {\em Proceedings of the Genetic and Evolutionary Computation
  Conference (GECCO2015)}, pages 567--573.

\bibitem[Mnih et~al., 2016]{Mnih2016}
Mnih, V., Badia, A.~P., Mirza, M., Graves, A., Lillicrap, T.~P., Harley, T.,
  Silver, D., and Kavukcuoglu, K. (2016).
\newblock Asynchronous methods for deep reinforcement learning.
\newblock In {\em Proceedings of the International Conference on Machine
  Learning (ICML2016)}, pages 1928--1937.

\bibitem[Mnih et~al., 2015]{Mnih15}
Mnih, V., Kavukcuoglu, K., Silver, D., Rusu, A.~A., Veness, J., Bellemare,
  M.~G., Graves, A., Riedmiller, M., Fidjeland, A.~K., Ostrovski, G., Petersen,
  S., Beattie, C., Sadik, A., Antonoglou, I., King, H., Kumaran, D., Wierstra,
  D., Legg, S., and Hassabis, D. (2015).
\newblock Human-level control through deep reinforcement learning.
\newblock {\em Nature}, 518(7540):529--533.

\bibitem[Nair et~al., 2015]{Nair15}
Nair, A., Srinivasan, P., Blackwell, S., Alcicek, C., Fearon, R., Maria, A.~D.,
  Panneershelvam, V., Suleyman, M., Beattie, C., Petersen, S., Legg, S., Mnih,
  V., Kavukcuoglu, K., and Silver, D. (2015).
\newblock Massively parallel methods for deep reinforcement learning.
\newblock {\em CoRR}, abs/1507.04296.

\bibitem[Rummery and Niranjan, 1994]{Rummery94}
Rummery, G.~A. and Niranjan, M. (1994).
\newblock On-line {Q}-learning using connectionist systems.
\newblock Technical Report CUED/F-INFENG/TR 166, Cambridge University
  Engineering Department.

\bibitem[Schaul et~al., 2016]{Schaul2016}
Schaul, T., Quan, J., Antonoglou, I., and Silver, D. (2016).
\newblock Prioritized experience replay.
\newblock In {\em International Conference on Learning Representations
  (ICLR2016)}.

\bibitem[Scherrer et~al., 2015]{Scherrer15}
Scherrer, B., Ghavamzadeh, M., Gabillon, V., Lesner, B., and Geist, M. (2015).
\newblock Approximate modified policy iteration and its application to the game
  of tetris.
\newblock {\em Journal of Machine Learning Research}, 16:1629--1676.

\bibitem[Smolensky, 1986]{Smolensky86}
Smolensky, P. (1986).
\newblock Information processing in dynamical systems: Foundations of harmony
  theory.
\newblock In Rumelhart, D.~E. and McClelland, J.~L., editors, {\em Parallel
  Distributed Processing: Explorations in the Microstructure of Cognition.
  Volume 1: Foundations}. MIT Press.

\bibitem[Sutton, 1988]{Sutton88}
Sutton, R.~S. (1988).
\newblock Learning to predict by the method of temporal differences.
\newblock {\em Machine Learning}, 3:9--44.

\bibitem[Sutton, 1996]{Sutton96}
Sutton, R.~S. (1996).
\newblock Generalization in reinforcement learning: Successful examples using
  sparse coarse coding.
\newblock In {\em Proceedings of Advances in Neural Information Processing
  Systems (NIPS1996)}, pages 1038--1044.

\bibitem[Sutton and Barto, 1998]{Sutton98}
Sutton, R.~S. and Barto, A. (1998).
\newblock {\em Reinforcement Learning: An Introduction}.
\newblock MIT Press.

\bibitem[Szita and Szepesv{\'{a}}ri, 2010]{Szita10}
Szita, I. and Szepesv{\'{a}}ri, C. (2010).
\newblock {SZ-T}etris as a benchmark for studying key problems of reinforcement
  learning.
\newblock In {\em ICML 2010 workshop on machine learning and games}.

\bibitem[Tesauro, 1994]{Tesauro1994}
Tesauro, G. (1994).
\newblock Td-gammon, a self-teaching backgammon program, achieves master-level
  play.
\newblock {\em Neural Computation}, 6(2):215--219.

\bibitem[Thiery and Scherrer, 2009]{Thiery09}
Thiery, C. and Scherrer, B. (2009).
\newblock Improvements on learning tetris with cross entropy.
\newblock {\em International Computer Games Association Journal}, 32.

\bibitem[Thrun and Schwartz, 1993]{Thrun93}
Thrun, S. and Schwartz, A. (1993).
\newblock Issues in using function approximation for reinforcement learning.
\newblock In {\em Proceedings of the 1993 Connectionist Models Summer School},
  pages 255--263.

\bibitem[van Hasselt, 2010]{Hasselt10}
van Hasselt, H. (2010).
\newblock Double q-learning.
\newblock In {\em Proceedings of Advances in Neural Information Processing
  Systems (NIPS2010)}, pages 2613--2621.

\bibitem[van Hasselt et~al., 2015]{Hasselt15}
van Hasselt, H., Guez, A., and Silver, D. (2015).
\newblock Deep reinforcement learning with double q-learning.
\newblock {\em CoRR}, abs/1509.06461.

\bibitem[Wang et~al., 2016]{Wang2016}
Wang, Z., Schaul, T., Hessel, M., van Hasselt, H., Lanctot, M., and de~Freitas,
  N. (2016).
\newblock Dueling network architectures for deep reinforcement learning.
\newblock In {\em Proceedings of the International Conference on Machine
  Learning (ICML2016)}, pages 1995--2003.

\end{thebibliography}
\bibliographystyle{apalike}

\end{document}